%% file: main.tex
\documentclass{article}

\PassOptionsToPackage{numbers, compress}{natbib}

\usepackage[preprint]{neurips_2021}

\usepackage[utf8]{inputenc} 
\usepackage[T1]{fontenc}    
\usepackage{hyperref}       
\usepackage{url}            
\usepackage{booktabs}       
\usepackage{amsfonts}       
\usepackage{nicefrac}       
\usepackage{microtype}      
\usepackage{xcolor}         
\usepackage{graphicx}
\usepackage{mathtools}
\usepackage{listliketab}
\usepackage[symbol]{footmisc}
\usepackage{comment}
\usepackage{enumitem}
\RequirePackage{authblk}

\usepackage{placeins}
\usepackage{float}

\definecolor{MyDarkBlue}{rgb}{0,0.08,1}
\definecolor{MyDarkGreen}{rgb}{0.02,0.6,0.02}
\definecolor{MyDarkRed}{rgb}{0.8,0.02,0.02}
\definecolor{MyDarkOrange}{rgb}{0.40,0.2,0.02}
\definecolor{MyPurple}{RGB}{111,0,255}
\definecolor{MyRed}{rgb}{1.0,0.0,0.0}
\definecolor{MyGold}{rgb}{0.75,0.6,0.12}
\definecolor{MyDarkgray}{rgb}{0.66, 0.66, 0.66}

\title{Physion: Evaluating Physical Prediction from Vision in Humans and Machines}

\author[ ]{Daniel M. Bear$^{1, 4,}$\footnote[1]{}}
\author[ ]{Elias Wang$^{2, 4,}$\footnote[1]{}}
\author[ ]{Damian Mrowca$^{3,}$\footnote[1]{}}
\author[ ]{Felix Binder$^{5,}$\footnote[1]{}}
\author[1,7]{Hsiao-Yu Fish Tung}
\author[7]{R.T. Pramod}
\author[6]{Cameron Holdaway}
\author[6]{Sirui Tao}
\author[7]{Kevin Smith}
\author[3]{Fan-Yun Sun}
\author[3]{Li Fei-Fei}
\author[7]{Nancy~Kanwisher}
\author[7]{Joshua B. Tenenbaum}
\author[ ]{Daniel L. K. Yamins$^{1,3,4,*}$\footnote[1]{}}
\author[ ]{Judith Fan$^{6,*}$\footnote[1]{}}

\affil[ ]{Department of Psychology$^1$, Electrical Engineering$^2$, and Computer Science$^3$, and Wu~Tsai~Neurosciences~Institute$^4$, Stanford, CA 94305}
\affil[ ]{Department of Cognitive Science$^5$, and Psychology$^6$, UC San Diego, CA 92093}
\affil[ ]{Department of Brain and Cognitive Sciences and CBMM$^7$, MIT, Cambridge, MA 02139}
\affil[ ]{\newline \texttt{\{\href{mailto:dbear@stanford.edu}{dbear}, \href{mailto:eliwang@stanford.edu}{eliwang}, \href{mailto:mrowca@stanford.edu}{mrowca}\}@stanford.edu,
\href{mailto:fbinder@ucsd.edu}{fbinder@ucsd.edu}}}

\begin{document}

\footnotetext[1]{$^{/**}$Equal contribution}

\maketitle

\input{Sections/abstract_CR}
\input{Sections/introduction_CR}
\input{Sections/method_CR}
\input{Sections/experiments_CR}
\input{Sections/acknowledgments}
\input{Sections/broaderimpact}

\typeout{}
{\small

\input{main.bbl}
}
\newpage

\appendix

\input{Sections/supplementary}

\newpage
\end{document}

%% file: Sections/abstract_CR.tex

\vspace{-.3cm}
\begin{abstract}
While current vision algorithms excel at many challenging tasks, it is unclear how well they understand the physical dynamics of real-world environments.
Here we introduce \textbf{Physion}, a dataset and benchmark for rigorously evaluating the ability to predict how physical scenarios will evolve over time.
Our dataset features realistic simulations of a wide range of physical phenomena, including rigid and soft-body collisions, stable multi-object configurations, rolling, sliding, and projectile motion, thus providing a more comprehensive challenge than previous benchmarks.
We used \textbf{Physion} to benchmark a suite of models varying in their architecture, learning objective, input-output structure, and training data. 
In parallel, we obtained precise measurements of human prediction behavior on the same set of scenarios, allowing us to directly evaluate how well any model could approximate human behavior.
We found that vision algorithms that learn object-centric representations generally outperform those that do not, yet still fall far short of human performance.
On the other hand, graph neural networks with direct access to physical state information both perform substantially better and make predictions that are more similar to those made by humans. 
These results suggest that extracting physical representations of scenes is the main bottleneck to achieving human-level and human-like physical understanding in vision algorithms.
We have publicly released all data and code to facilitate the use of \textbf{Physion} to benchmark additional models in a fully reproducible manner, enabling systematic evaluation of progress towards vision algorithms that understand physical environments as robustly as people do. 
\end{abstract}

%% file: Sections/introduction_CR.tex
\section{Introduction}



Vision algorithms that understand the physical dynamics of real-world environments are key to progress in AI.
In many settings, it is critical to be able to anticipate when an object is about to roll into the road, fall off the table, or collapse under excess weight.
Moreover, for robots and other autonomous systems to interact safely and effectively with their environment they must be able to accurately predict the physical consequences of their actions. 

\subsection{Establishing Common Standards for Evaluating Physical Understanding}
Despite recent progress in computer vision and machine learning, it remains unclear whether any vision algorithms meet this bar of everyday physical understanding.
This is because previously developed algorithms have been evaluated against disparate standards ---
some prioritizing accurate prediction of every detail of a scenario's dynamics and others that only require predictions about a specific type of event. 



The first set of standards has generally been used to evaluate algorithms that operate on unstructured video inputs, such as in robotics \citep{ebert2018robustness}. 
These algorithms typically aim for fine-grained prediction of upcoming video frames or simulation of the trajectories of individual particles.
However, only algorithms with near-perfect knowledge of the world's physical state -- like Laplace's Demon -- could hope to predict how a complete set of events will unfold.
This explains why models of this kind have sufficed in less varied visual environments, but underfit on more diverse scenarios \citep{dasari2019robonet, mandlekar2018roboturk}.
Though recent efforts to scale these algorithms have led to improvements in the quality of predicted video outputs \citep{villegas2019high,wu2021greedy}, it remains to be seen whether their learned representations embody more general \textit{physical} knowledge.

The second set of standards has been used to probe qualitative understanding of physical concepts, especially in cognitive and developmental psychology \citep{baillargeon1985object, spelke1990principles, carey2001infants}.
Much of this work has focused on measuring and modeling human judgments about discrete events, such as whether a tower of blocks will fall over or whether an object will reemerge from behind an occluder \citep{battaglia2013simulation,bates2018modeling, baillargeon2011infants}.
Findings from this literature suggest that humans simulate dynamics over more abstract representations of visual scenes to generate reliable predictions at the relevant level of granularity \citep{riochet2018intphys, smith2019modeling}.
However, existing models that instantiate such simulations typically require require structured input data (e.g., object segmentations) that may not be readily available in real-world situations \citep{lerer2016physnet,janner2018op2}.
Moreover, the abstractions that are appropriate for one task may not work well in more general settings \citep{veerapaneni2020entity, watters2019cobra, nair2020goalaware}.

A key challenge in developing improved visual models of physical understanding is thus to establish common standards by which to evaluate them. 
Here we propose such a standard that both combines elements of previous approaches and goes beyond them: we require models to operate on highly varied and unstructured visual inputs to generate event-based predictions about a wide variety of physical phenomena.
By contrast with prior efforts to evaluate vision algorithms, our proposed standard argues for the importance of considering a wider variety of physical scenarios and the ability to compare model predictions directly with human judgments. 
By contrast with prior efforts to model human physical understanding, our approach embraces the challenge of generating predictions about key events from realistic visual inputs.  


\subsection{Desiderata for a Generalized Physical Understanding Benchmark}

We envision our generalized physical understanding benchmark as combining two key components: first, a dataset containing visually realistic and varied examples of a wide variety of physical phenomena; and second, a generic evaluation protocol that probes physical understanding in a way that is agnostic to model architecture and training regime.
 
\paragraph{Dataset.} While there are several existing datasets that probe physical understanding to some extent, each of them fall short on at least one key dimension. 
Some datasets contain realistic visual scenes but do not adequately probe understanding of object dynamics \cite{dasari2019robonet}.
Other datasets feature realistic scenarios with challenging object dynamics, but consider only a narrow set of physical phenomena, such as whether a tower of blocks will fall \cite{groth2018shapestacks} or whether a viewed object's trajectory violates basic physical laws \cite{riochet2018intphys, piloto2018probing, smith2019modeling}. 
Other datasets featuring a greater diversity of physical phenomena are designed in simplified 2D environments that may not generalize to real-world 3D environments \cite{bakhtin2019phyre}.

\paragraph{Evaluation protocol.} In order to test a wide variety of models in a consistent manner, many commonly used evaluations will not suffice. 
For example, evaluations that query the exact trajectories of specific objects \citep{battaglia2016interaction,chang2016compositional} are not well posed for models that do not extract explicit object representations. 
Conversely, evaluations that depend on image matching or visual realism-based metrics  \cite{finn2016unsupervised,ye2019compositional, dasari2019robonet, wu2021greedy} are not straightforward to apply to models that do not re-render images.
A more promising approach to measuring physical understanding in a model-agnostic manner may instead take inspiration from prior work investigating human physical prediction ability \cite{battaglia2013simulation, sanborn2013reconciling, bates2018modeling}, which does not assume that the trajectories of all objects in a scene are represented with perfect fidelity.


\subsection{Physion: A Dataset and Benchmark for Physical Understanding}

\begin{figure}[t]
\vspace{-0.7cm}
\centering
    \includegraphics[width=0.9\textwidth]{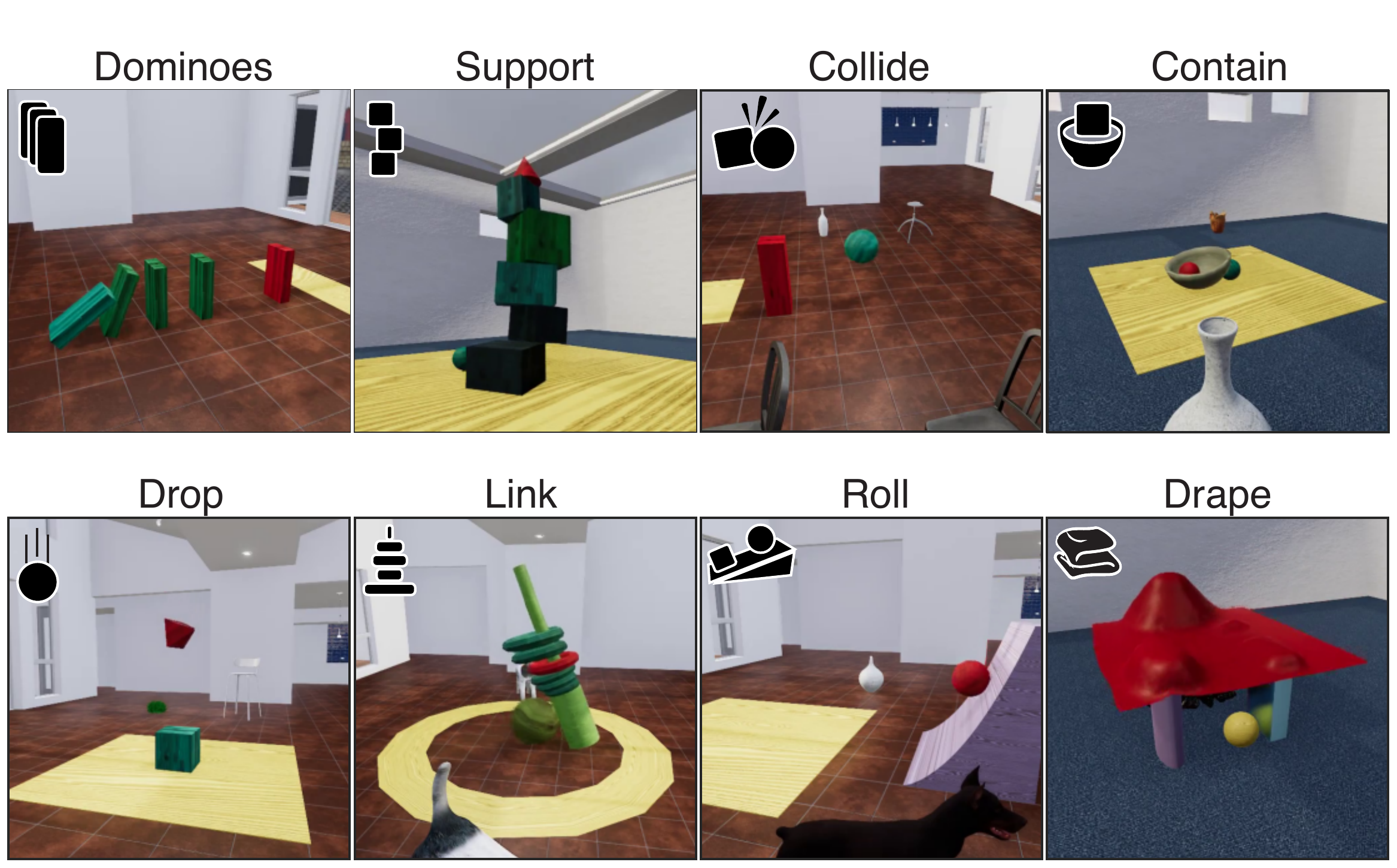}
\caption{Example frames from the eight \textbf{Physion} scenarios. Red object is agent; yellow is patient.}
\label{fig:scenario_display}
\vspace{-0.3cm}
\end{figure}

In recognition of the above desiderata, we developed \textbf{Physion}, a new physical understanding dataset and benchmark. 
Our dataset contains a wide variety of visually realistic examples of familiar physical phenomena, including:
collisions between multiple objects; 
object-object interactions such as support, containment, and attachment;
projectile, rolling, and sliding motion that depends on object geometry;
and the behavior of soft materials like cloth.
For each of these eight scenario types (\ref{fig:scenario_display}), we operationalize physical understanding using the \textbf{object contact prediction (OCP) task}, which prompts agents to predict whether two cued objects will come into contact as a scene unfolds.

\subsection{Using Physion to Benchmark Human and Model Physical Understanding}

In addition to the dataset, we introduce a unified evaluation protocol for directly comparing model and human behavior.
Approximating human physical understanding from vision is a natural target for AI systems for two key reasons: first, humans have already demonstrated their ability to competently navigate a wide variety of real-world physical environments; and second, it is important for AI systems to anticipate how humans understand their physical surroundings in order to co-exist safely with people in these environments. 
Towards this end, our paper conducts systematic comparison between humans and several state-of-the-art models on the same physical scenarios. 


Our experiments feature a wide range of models that vary in their architecture, learning objective, input-output structure, and training regime.
Specifically, we include vision models that make pixel-level predictions via fully convolutional architectures, \citep{fragkiadaki2015learning, agrawal2016learning, li2016fall, finn2016unsupervised, lerer2016physnet, zhang2016comparative, mottaghi2016newtonian, mottaghi2016happens, vondrick2016anticipating, haber2018learning, lee2018stochastic, schmeckpeper2019learning, girdhar2020forward}; 
those that either explicitly learn object-centric representations of scenes \citep{veerapaneni2020entity, kipf2019contrastive, ding2020object, girdhar2020forward, riochet2020occlusion} or are encouraged to learn about objects via supervised training \citep{simonyan2014vgg, touvron2020deit};
and physics dynamics models that operate on object- or particle-graph representations provided as input \citep{chang2016compositional, battaglia2016interaction, li2018learning, bates2018modeling, tacchetti2018relational, sanchez2018graph, battaglia2018relational, mrowca2018flexible, ajay2019combining, smith2019modeling, ye2019compositional, qi2020learning}.

Models that perform physical simulation on a graph-like latent state are especially attractive candidates for approximating human prediction behavior, based on prior work that has found that \textit{non-}machine learning algorithms that add noise to a hard-coded simulator accurately capture human judgments in several different physical scenarios \citep{battaglia2013simulation, sanborn2013reconciling, bates2015humans, bi2021perception}.
Consistent with these results, recurrent graph neural networks supervised on physical simulator states can learn to accurately predict full object trajectories \citep{mrowca2018flexible, li2018learning, li2020visual, sanchez2020learning}.
However, these models have not been tested for their ability to generalize across diverse, multi-object scenarios, and they require such detailed physical input and trajectory supervision that they have so far not been useful in cases where only realistic sensory observations are available.

Among models that take visual input, object-centric predictors in some cases make more accurate predictions than those that simulate scene dynamics in pixel space \citep{veerapaneni2020entity, qi2020learning, ding2020object};
however, these comparisons have only been done in reduced environments with few distinct physical phenomena, so it is not known whether this result holds in more realistic settings.
Indeed, models that make pixel-level predictions are standard in robotics applications \citep{lee2018stochastic, wu2021greedy} due to the longstanding difficulty of inferring accurate object-centric representations from raw video data without supervision, despite recent progress \citep{burgess2019monet, veerapaneni2020entity, bear2020learning}.

\subsection{Summary of Key Findings}
By assessing many models on the same challenging physical understanding task, our experiments address previously unresolved questions concerning the roles of model architecture, dataset, and training protocols in achieving robust and human-like physical understanding. 
We found that no current vision algorithms achieve human-level performance in predicting the outcomes of \textbf{Physion} scenes.
Vision algorithms encouraged to learn object-centric representations generally outperform those that do not, yet still fall far short of human performance.
On the other hand, particle-based models with direct access to physical state information both perform substantially better and make predictions that are more similar to those made by humans. 
Taken together, these results suggest that extracting physical representations of visual scenes is the key bottleneck to achieving human-level and human-like physical understanding in vision algorithms.


\subsection{Our Vision for Physion}
Our initial public release of \textbf{Physion} includes large, labeled training and test datasets for each scenario, as well as code for for generating additional training data.
As such, one potential way to use \textbf{Physion} is to train additional models directly on the OCP task for one or more of the scenarios, yielding, for example, a model that excels at predicting whether block towers will fall.
However, the primary use case we have in mind for \textbf{Physion} is to test how well pretrained models transfer to challenging physical understanding tasks, analogous to how humans make predictions about \textbf{Physion} videos without extensive training on the OCP task.
Towards this end, we have shared code to facilitate the use of the \textbf{Physion} test dataset to benchmark additional models in a fully reproducible manner, enabling systematic evaluation of progress towards vision algorithms that understand physical environments as robustly as people do. 

%% file: Sections/method_CR.tex
\section{Methods}
\subsection{Benchmark Design}

We used the ThreeDWorld simulator (TDW), a Unity3D-based environment \citep{gan2020threedworld}, to create eight physical scenarios out of simple objects that incorporate diverse physical phenomena (Fig.~\ref{fig:scenario_display}):

\begin{listliketab} 
\storestyleof{itemize} 
\begin{tabular}{@{\hskip-2pt}l@{\hskip6pt}l@{\hskip5pt}l@{\hskip5pt}l}
1. & \textbf{Dominoes} & -- & sequences of collisions that depend on the arrangement and poses of objects \\
2. & \textbf{Support} & -- & stacks of objects that may fall over, depending on their shapes and arrangement \\
3. & \textbf{Collide} & -- & pairs of objects that may collide, depending on their placement and trajectories \\
4. & \textbf{Contain} & -- & container-like objects that may constrain other objects by virtue of their shapes \\
5. & \textbf{Drop} & -- & objects falling and bouncing under the force of gravity \\
6. & \textbf{Link}  & -- & objects restricted in their motion because they are attached to other objects \\
7. & \textbf{Roll} & -- & objects that move across a surface either by rolling or sliding \\
8. & \textbf{Drape} & -- & cloth draping over other objects by virtue of their shape and the cloth's material. \\
\end{tabular} 
\end{listliketab}

In each scenario, contact between agent and patient serves as a non-verbal indicator of some physical higher-order variable -- whether a tower fell over, a bowl contained a ball, a torus was attached to a post -- whose prediction should require understanding of the relevant physical phenomena.
Together, these scenarios cover much of the space of physical dynamics possible through simple rigid- and soft-body interactions; additional scenarios will be developed to include other material types (e.g., ``squishy'' objects, fluids) and complex interactions (e.g. multi-part, jointed objects.)

\begin{figure}[t]
\centering
    \includegraphics[width=0.8\textwidth]{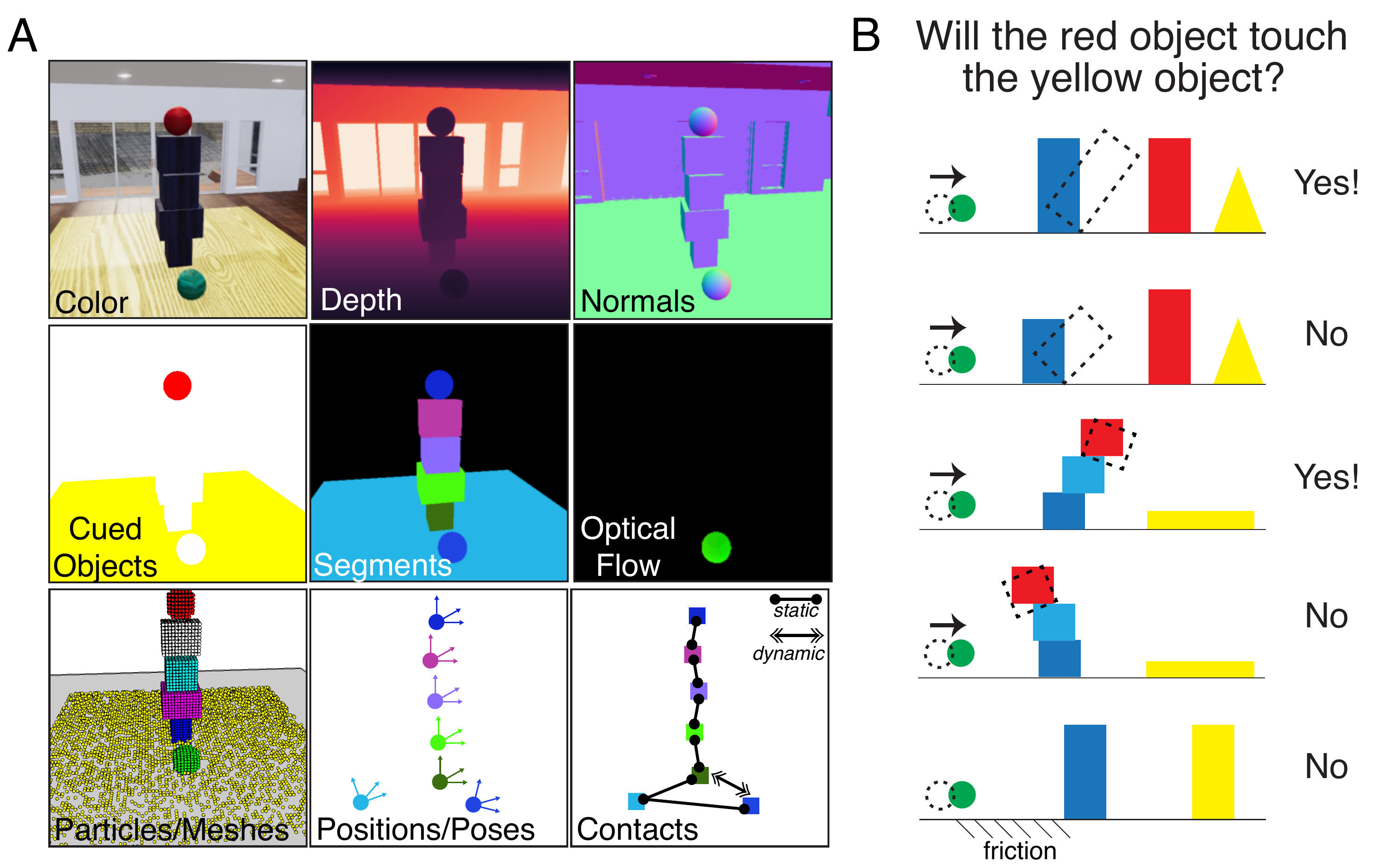}
\caption{Stimulus attributes and task design. (\textbf{A}) Output of TDW for an example frame of a stimulus movie. (\textbf{B}) A schematic of the OCP task: humans and models must predict whether the \textit{agent} object (red) will contact the \textit{patient} (yellow), given the initial setup and the motion of the \textit{probe} (green).}
\label{fig:tdw_data}
\vspace{-0.4cm}
\end{figure}

\subsection{Stimulus Generation and Task Design}
We constructed scenes out of basic ``toy blocks'' to avoid confounds from knowledge of object configurations that are common in the real world (e.g., cups typically appearing on tables); 
rather, accurate predictions should require judgments about objects' physical properties, relationships, and dynamics.
To increase physical variability \textit{within} each scenario, we identified multiple configurations of simulator parameters that lead to different types of physical dynamics.
Configurations specify distributions of initial scene variables, such as the positions of objects;
they also introduce substantial visual variation that does not affect the physical outcome of the scene, including variation in camera position and pose, object colors and textures, the choice of ``distractor'' object models that do not participate in scene dynamics, and the appearance of the background.
Training and testing stimuli were generated by randomly sampling initial conditions and scene properties according to each configuration, then running the simulation until all objects came to rest.
Additional stimuli can be generated by sampling further from our configurations or by creating new ones.
Examples of stimuli from each scenario can be found in the Supplement.

Each stimulus is a 5-10 second movie rendered at 30 frames per second. 
For model training and evaluation we also supply the full output of the TDW simulation (Fig. \ref{fig:tdw_data}A), which includes:
1.) \emph{visual data per frame}: color image, depth map, surface normal vector map, object segmentation mask, and optical flow map;
2.) \emph{physical state data per frame}: object centroids, poses, velocities, surface meshes (which can be converted to particles), and the locations and normal vectors for object-object or object-environment collisions;
3.) \emph{stimulus-level labels and metadata}: the model names, scales, and colors of each object; the intrinsic and extrinsic camera matrices; segmentation masks for the agent and patient object and object contact indicators; the times and vectors of any externally applied forces; and scenario-specific parameters, such as the number of blocks in a tower.
All stimuli from all eight scenarios share a common OCP task structure (Fig. \ref{fig:tdw_data}B):
there is always one object designated the \textit{agent} and one object designated the \textit{patient}, and most scenes have a \textit{probe} object whose initial motion sets off a chain of physical events.
Models and people are asked to predict whether the agent and patient object will come into contact by the time all objects come to rest.
We generated trials for human testing by sampling from scenario-specific configurations until we had 150 testing stimuli per scenario with an equal proportion of contact and no-contact outcomes.

\subsection{Testing Humans on the Physics Prediction Benchmark}
\paragraph{Participants.}
800 participants (100 per scenario; 447 female, 343 male, 7 declined to state; all native English speakers) were recruited from Prolific and paid \$4.00 for their participation. Each was shown all 150 stimuli from a single scenario.
Data from 112 participants were excluded for not meeting our preregistered inclusion criterion for accurate and consistent responses on attention-check trials (see Supplement).
Our preregistered analysis plan is stored under version control in our GitHub repository.
These studies were conducted in accordance with the UC San Diego and Stanford IRBs.

\begin{figure}[ht]
    \centering
    \includegraphics[width=\textwidth]{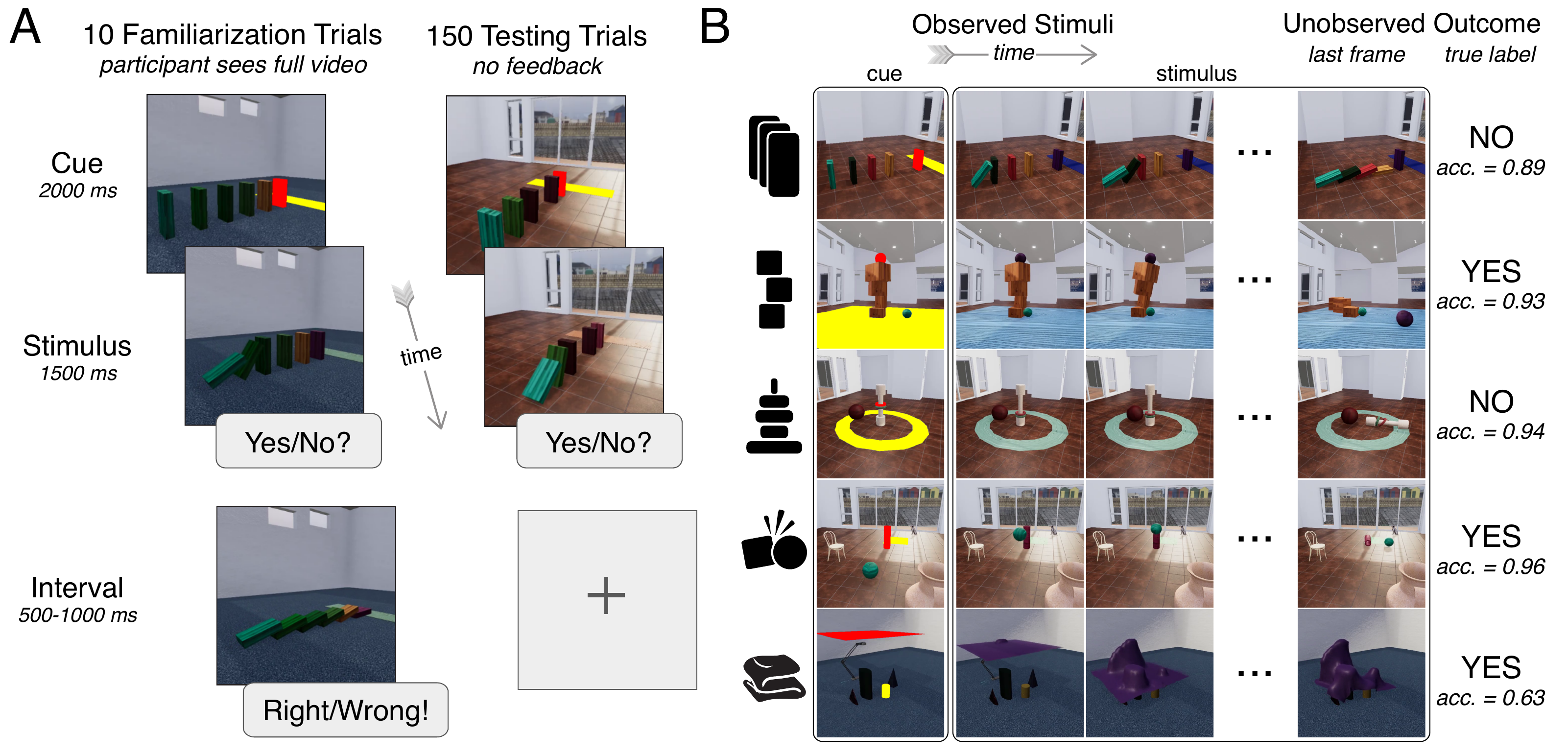}
    \caption{Human task. (A) Trial structure for the familiarization trials (\textit{left}) and test trials (\textit{right}) indicating the Cue, Stimulus, and Inter-trial periods. (B) Example stimuli (rows) including the last frame (not shown during the experiment). Last column indicates the outcome and human accuracy.}
    \label{fig:experimental_design}
    \vspace{-0.2cm}
\end{figure}

\paragraph{Task procedure.}
The structure of our task is shown in Fig. \ref{fig:experimental_design}A.
Each trial began with a fixation cross, which was shown for a randomly sampled time between 500ms and 1500ms.
To indicate which of the objects shown was the agent and patient object, participants were then shown the first frame of the video for 2000ms. During this time, the agent and patient objects were overlaid in red and yellow respectively. The overlay flashed on and off with a frequency of 2Hz. 
After this, the first 1500ms of the stimulus were played. After 1500ms, the stimulus was removed and the response buttons were enabled. Participants proceeded to the next trial after they made a prediction by selecting either ``YES'' (the agent and patient would touch) or ``NO'' (they would not). The order of the buttons was randomized between participants. 
Before the main task, participants were familiarized with 10 trials that were presented similarly to the test trials, except (a) the full stimulus movie and accuracy feedback was presented after participants indicated their prediction, and (b) all trials were created from basic templates without occluding and distracting objects. Familiarization trials were always presented in the same order.
After the test trials were completed, basic demographics were collected from participants. Finally, participants were informed of their overall accuracy. 

\subsection{Benchmarking Computer Vision and Physical Dynamics Models}
\begin{figure}[t]
\centering
    \includegraphics[width=\textwidth]{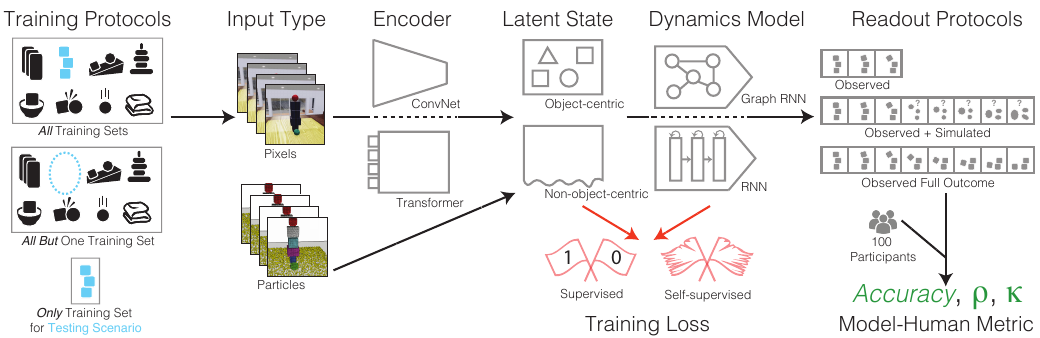}
\caption{The model benchmarking pipeline including training, architecture, and readout variants.}
\label{fig:modeling}
\vspace{-0.3cm}
\end{figure}

We developed a standard procedure for training machine learning models and evaluating any image- or physical state-computable algorithm on the benchmark.
Let $\{X_t\}_{i=1}^{N_{test}}$ be the set of $N_{test}$ testing stimuli for a single benchmark scenario, where $\{X_t\}_i$ denotes the ordered set of RGB images that constitutes the full movie of stimulus $i$ and $\{X_{1:t_{vis}}\}$ the truncated movie shown to participants.
Further let $\mathcal{O}_i := \{o_1,o_2,...,o_{K_i}\}$ denote unique IDs for each of the $K$ objects being simulated in this stimulus.
Doing the OCP task can be formalized as making a binary contact prediction by applying to the testing stimuli a function $\mathcal{F}_{\Theta}: (\{X_{1:t_{vis}}\}, o_a, o_p) \mapsto P(contact)$, where $o_a$ is the agent object, $o_p$ is the patient, and $P(contact)$ is the predicted probability that they will come into contact.
For people, feedback on only ten familiarization trials is sufficient to learn such a function.
To adapt any image-computable model to the OCP task, we apply the following procedure.
First, we assume that a model can be decomposed into a \textit{visual encoder} that maps an input movie to a state-vector representation of each frame;
a \textit{dynamics predictor} that predicts unseen future states from the ``observed'' state vector; 
and a \textit{task adaptor} that produces a trial-level response $P(contact)$ from the concatenation of the observed and predicted state vectors (Fig. \ref{fig:modeling}).
In general, models will include only a visual encoder and possibly a dynamics predictor in their original design; 
the task adaptor is added and fit as part of our model evaluation pipeline, where it removes the need for the explicit trial-level cueing with superimposed object masks (see below.)

\textbf{Testing, Readout Fitting, and Training sets.}
Each \textbf{Physion} scenario consists of three stimulus sets: \textit{Testing}, \textit{Readout Fitting}, and \textit{Training}.
The \textit{Testing} stimuli are identical to the 150 trials per scenario shown to humans, except that the agent and patient objects are permanently colored red and yellow (Fig. \ref{fig:scenario_display}) instead of being indicated by red and yellow masks on the first frame (Fig. \ref{fig:experimental_design}).
This difference allows models to be tested on RGB movie stimuli alone, without providing segmentation masks that most computer vision model architectures are not designed to handle as inputs.
Each trial in the \textit{Testing} sets includes the ground truth label of whether it ends in agent-patient contact and the responses of >100 human participants.
We also provide the \textit{Human Testing} stimuli with red and yellow cueing masks rather than permanently colored objects.

Each scenario's \textit{Readout Fitting} set consists of 1000 stimuli generated from the same configurations as the \textit{Testing} stimuli, such that the two sets have the same visual and physical statistics.
The \textit{Readout Fitting} stimuli are for fitting a OCP task-specific adaptor to each model. 
In designing \textbf{Physion}, we did not want to restrict testing only to models optimized directly to do the OCP prediction task.
Thus, during evaluation we freeze the parameters of a pretrained model and fit a generalized linear model, the task adaptor, on various subsets of model features (see below).
The \textit{Readout Fitting} stimuli are the training set for this fitting procedure, with the ground truth object contact labels acting as supervision.
This allows the task adaptor to generalize to the \textit{Testing} stimuli.

Finally, each scenario's \textit{Training} set includes 2000 movies generated from the same configurations as the \textit{Testing} and \textit{Readout Fitting} stimuli, but with no visual features indicating agent and patient objects.
The purpose of the \textit{Training} sets is to let models learn or fine-tune representations of physical dynamics in a way that is agnostic to any particular readout task:
a model partly or entirely trained on a ``non-physics'' task like object categorization might nevertheless acquire a human-like representation of the physical world, which \textbf{Physion} should reveal via transfer learning.
During training models see movie clips sampled from the entirety of each \textit{Training} stimulus, not just the initial portion seen during readout fitting and testing, and they do not receive ground truth OCP labels.

The procedure for training a given model depends on its original architecture and optimization procedure.
For models that take multi-frame inputs and include both a visual encoder and a dynamics predictor in their architecture, we train the full model end-to-end on the \textit{Training} sets.
For models that include only a visual encoder pretrained on another dataset and task (such as ImageNet), we add an RNN dynamics model that predicts future encoder outputs from the ``observed'' encoder outputs on an input frame sequence;
the training loss is the mean squared error between each predicted output and the matching observed output, which optimizes the dynamics model.
For these models, we train two versions: one in which the pretrained encoder parameters are fine-tuned and one in which they are frozen.
See \textbf{Model Comparison} below and the Supplement for further details.

\textbf{Model comparison.}
To get an overview of how current physical prediction algorithms compare to humans, we tested models from four classes (see Supplement for model details):
\begin{enumerate}[leftmargin=*]
    \item fully unsupervised, joint encoder-dynamics predictors trained only on the benchmark scenario data: \textbf{SVG} \citep{denton2018svg}, \textbf{OP3} \citep{veerapaneni2020entity}, \textbf{CSWM} \citep{kipf2019contrastive};
    \item encoder-dynamics models supervised on ground truth object data: \textbf{RPIN} \citep{qi2020learning};
    \item visual encoders \textit{pretrained with supervision on ImageNet} and extended with RNN dynamics predictors, which are trained in an \textit{unsupervised} way on the benchmark scenario data: \textbf{pVGG-mlp/lstm} \citep{simonyan2014vgg}, \textbf{pDeIT-mlp/lstm} \citep{touvron2020deit};
    \item particle-relation graph neural network dynamics predictors that take the ground truth simulator state as input and have no visual encoder (i.e. assume perfect observability of physical dynamics): \textbf{GNS} \citep{sanchez2020learning}, \textbf{GNS-RANSAC}, \textbf{DPI} \citep{li2018learning}.
\end{enumerate}

\textbf{Training protocols.}
We tested models given three types of training (Fig. \ref{fig:modeling}, left): 
\textit{all}, training on all scenarios' training sets concurrently; \textit{all-but}, training on all scenarios except the one the model would be tested on;
and \textbf{only}, training on only the scenario type the model would be tested on.
We consider the \textit{all} protocol to be the best test of physical understanding, since it produces a model that is not specialized to a specific scenario.
Differences between \textit{all} and \textit{all-but} or \textit{only} indicate how well a model can generalize across scenarios or overfit to a single scenario, respectively.

\paragraph{Testing protocols.} 
We fit logistic regression models as OCP task adaptors with three protocols (Fig. \ref{fig:modeling}, right): 
\textit{observed}, in which adaptors are fit only to the features produced by showing the human stimulus (first $t_{vis}$ frames, equivalent to 1.5 seconds) to the model's visual encoder;
\textit{observed+simulated}, which uses the \textit{observed} features concatenated with the ``simulated'' features output by the model's dynamics predictor;
and \textit{full}, which uses the features produced from showing the entire movie (not just the testing stimulus portion) to the visual encoder.
Outputs from the \textit{full} protocol cannot be directly compared to human data, since they represent a model's performance on a detection (rather than prediction) task; 
however, we use them to assess how well physical information is encoded in a model's visual features (see Experiments.)
We compare a model's outputs to human responses on each scenario's testing stimuli with three standard metrics (Fig. \ref{fig:modeling}, right): overall accuracy, Pearson correlation between model and average human responses across stimuli, and Cohen's $\kappa$, a measure of how much a model's binary predictions resemble a single human's, averaged across participants.
For all three metrics, we assess how close models are to the ``human zone'' -- the empirical distribution of each statistic across humans or human-human pairs.

%% file: Sections/experiments_CR.tex
\section{Results and Discussion}

\begin{figure}[t]
    \centering
    \includegraphics[width=0.99\textwidth]{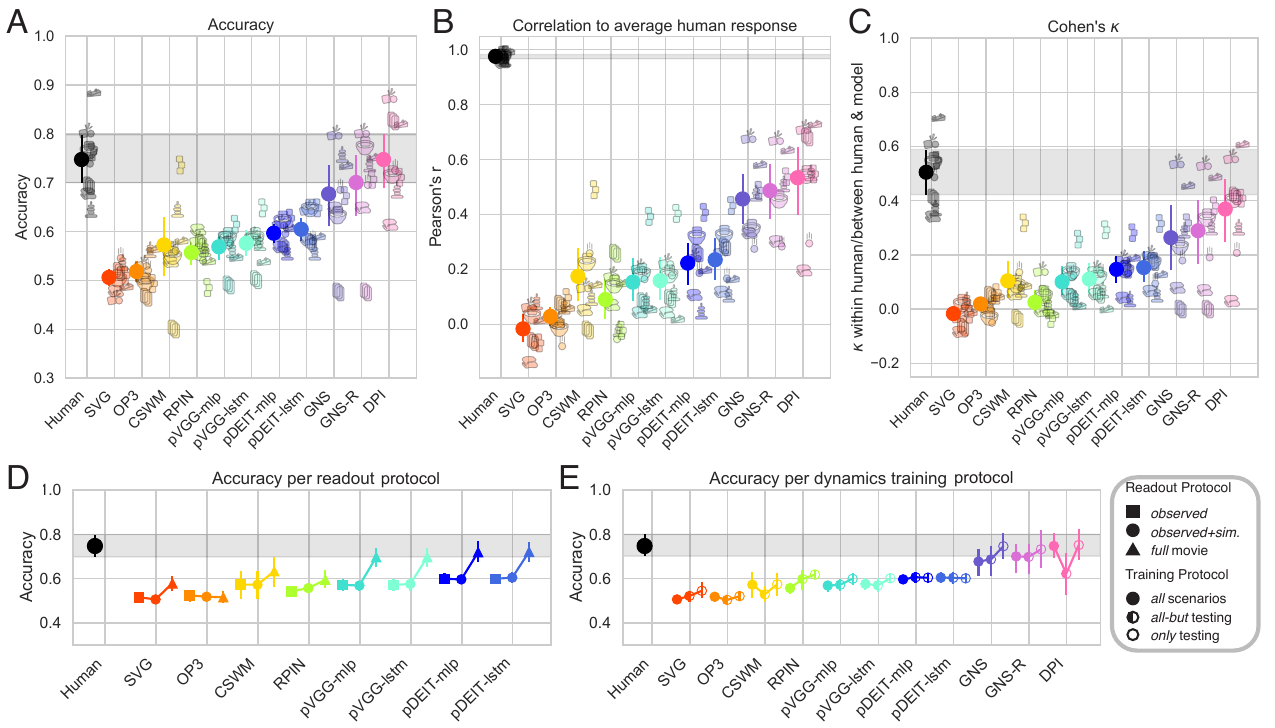}
    \caption{Comparisons between humans and models. First row: the \textit{all}-scenarios trained, \textit{observed+simulated}-readout task accuracy (\textbf{A}), Pearson correlation between model output and average human response (\textbf{B}), and Cohen's $\kappa$ (\textbf{C}) for each model on each scenario, indicated by its icon. Black icons and the gray zones (2.5th-97.5th percentile) show human performance, mean correlation between split halves of participants, and mean human-human Cohen's $\kappa$, respectively. Second row: accuracy of models across the three readout (\textbf{D}) and training (\textbf{E}) protocols; note that particle-input models have only the \textit{observed+simulated} readout protocol, as predictions are made based solely on whether two objects came within a threshold distance at the end of the predicted dynamics.}
    \label{fig:model_human_comparision}
    \vspace{-0.3cm}
\end{figure}

\textbf{Human behavior is reliable, with substantially above-chance performance.}
Human performance was substantially above chance across all eight scenarios (proportion correct = 0.71, t=27.5, p<$10^{-7}$, Fig. \ref{fig:model_human_comparision}A), though there was variation in performance across scenarios (e.g., higher accuracy on \textbf{Roll} than \textbf{Link} or \textbf{Drape}).
Moreover, the ``human zones'' for all metrics (raw performance, correlation-to-average, and Cohen's $\kappa$) were tight and far from chance (gray horizontal bars in Fig. \ref{fig:model_human_comparision}A-E), showing that the human response patterns were highly reliable at our data collection scale and thus provide a strong empirical test for discriminating between models.
Interestingly, each scenario included some stimuli on which the participant population scored significantly \textit{below} chance (Fig. \ref{fig:adversarial}).
Many of these ``adversarial'' stimuli had objects teetering on the brink of falling over or other unlikely events occurring after the observed portion of the movie.
People may have accurately judged that most scenes \textit{similar to} the observed stimulus would have one outcome, unaware that the other outcome actually occurred due to a physical fluke.
This pattern of reliable errors is especially useful for comparing models with humans: if stimuli that fool people do not fool a model, it would suggest that the model draws on different information or uses a non-human strategy for making predictions.

\textbf{Particle-based models approach human performance levels, with strong generalization.}
Models that received ground-truth TDW object particles as input and supervision (\textbf{GNS}, \textbf{GNS-RANSAC}, \textbf{DPI}) matched human accuracy on many scenarios, with the object-centric \textbf{DPI} reaching across-scenario human performance levels (Fig. \ref{fig:model_human_comparision}A).
These data are consistent with findings that probabilistic physical simulations can account for behavioral judgments on single scenarios that resemble ours \citep{battaglia2013simulation, sanborn2013reconciling, bates2015humans, bi2021perception}.
However, our results go beyond prior work in several ways.
First, these three models are graph neural networks that \textit{learn} to simulate physical scenes rather than assuming access to a ``noisy'' version of ground truth dynamics directly provided by the physics engine.
Second, the models here performed well above chance when trained with the \textit{all} and \textit{all-but} protocols, not just when they were fit to single scenario types (\textit{only}) as in the work where they were developed \citep{li2018learning, sanchez2020learning} (Fig. \ref{fig:model_human_comparision}A,E). 
These results imply that a single graph neural network can learn to make human-level physical predictions across a diverse set of physical scenarios.

\textbf{Vision-based models substantially underperform humans, but object-related training may help.}
Particle input models have an enormous advantage over both humans and vision models: they operate on ground truth physical information that, in the real world, can never be observed directly, such as the 3D positions, poses, trajectories, and fine-scale shapes of all objects and their occluded surfaces.
Whereas humans overcome these limits, none of the vision algorithms here came close to performing at human levels (Fig. \ref{fig:model_human_comparision}A).
Not all vision models were equally far off, though:
among those whose encoders and dynamics simulators were fully unsupervised, \textbf{SVG}, a model with only convolutional latent states, performed nearly at chance levels; \textbf{OP3}, an object-centric model trained by rendering pixel-level future predictions (b=0.06, t=7.6, p<$10^{-11}$), performed marginally better; while \textbf{CSWM}, a model with contrastively-learned object-centric latent states, significantly outperformed both \textbf{SVG} and \textbf{OP3}. 
Interestingly, the \textit{supervised} object-centric model \textbf{RPIN} was only more accurate than \textbf{CSWM} when trained with the \textit{all-but} and \textit{only} protocols, but not the \textit{all} protocol (b=0.035, t=3.7, p<$10^{-3}$, Fig. \ref{fig:model_human_comparision}A,E);
further experiments are needed to test whether exactly matching the architectures of the two models would reveal a larger effect of ground truth supervision.
Together, these results suggest that learning better object-centric representations from realistic, unlabeled video should be a core aim of visual prediction approaches.

The models with \textit{ImageNet-pretrained} ConvNet encoders (\textbf{pVGG-mlp/lstm}) significantly outperformed the best fully TDW-trained models (\textbf{CSWM}, \textbf{RPIN}, b=0.015, t=2.9, p<0.01), and were themselves outperformed by models with ImageNet-pretrained Transformer encoders (\textbf{pDeIT-mlp/lstm}, b=0.067, t=16.5, p<$10^{-15}$).
This suggests that (supervised) ImageNet pretraining and a better (and perhaps, more ``object-aware''-attention driven) encoder architecture produce visual features that are better for physical prediction even \textit{without} learning to explicitly simulate the future.
Together these results highlight the importance of learning a ``good'' visual representation; vision algorithms may benefit from training their encoders on separate tasks and data before learning dynamics predictors.

\textbf{Error-pattern consistency is strongly correlated with performance, but a substantial gap remains.}
A striking feature of our results is that error-pattern consistency as measured either by correlation-to-average human or Cohen's $\kappa$ (Fig. \ref{fig:model_human_comparision}B-C) is itself strongly correlated with absolute model performance. In other words, models that performed better on the prediction task also made errors that were more like those made by humans, strongly analogous to the situation with core visual object recognition~\cite{rajalingham2018large}. This result suggests, albeit weakly, that human behavior has been highly optimized either directly for a prediction task like that measured in this paper, or for something highly correlated with it. 
However, none of the models fully reached the ``human zone'' in which their outputs would be statistically indistinguishable from a person's.
This means that even the particle-based models can be improved to better match the judgments people make, including errors;
prior work suggests that adding noise to these models could better recapitulate human mental ``simulation'' \citep{battaglia2013simulation,bates2018modeling,smith2013sources}.
Consistent with this possibility, we found that the particle-based models' predictions were uncorrelated with human predictions on the ``adversarial'' stimuli, many of which would have opposite outcomes if their initial conditions were slightly different (Fig. \ref{fig:hardchanceeasy}). 
Adding noise to the models' forward dynamics might therefore mimic how humans make predictions about \textit{probable} outcomes, rather than simulating dynamics so precisely that they capture even rare flukes.

\textbf{What have vision-based models actually learned?}
Vision model predictions from the \textit{observed+simulated} readout protocol were, overall, no better than predictions from the \textit{observed} protocol (p=0.53, Fig. \ref{fig:model_human_comparision}D). 
This implies that none of the visual dynamics models learned to ``simulate'' anything about the scenes that helped on the OCP task (though dynamics predictions during end-to-end training could have usefully shaped the encoder representations.)
Rather, any above-chance performance for the vision models was likely due to having visual features that could discriminate some trial outcomes from cues in the initial movie segment.
Understanding what makes these visual features useful is the subject of ongoing work: they could be an example of non-causal ``shortcut learning'' \citep{geirhos2020shortcut} or they could encode important physical properties like object position, shape, and contact relationships.
The latter possibility is further supported by two observations. 
First, the \textit{full} readout protocol yielded significantly higher accuracy for the vision models (b=0.094, t=12.0, p<$10^{-15}$, Fig. \ref{fig:model_human_comparision}D), indicating that the learned visual features \emph{are} useful for object contact \textit{detection}. Thus, the best visual features carry some information about the observed objects' spatial relationships, and their relative failures in the \textit{observed} protocol can be fairly said to be these models' lack of physical ``understanding.''
Second, the ImageNet-pretrained models benefited the most from observing the full movie, raising the possibility that their pretraining actually captured \textit{more} physically-relevant information than object-centric learning on TDW.
Untangling this will require finer-scale comparison between encoder architectures, training datasets, and various supervised and self-supervised losses.

\textbf{Having sufficient variability across physical scenarios promotes strong generalization.}
Compared to models trained concurrently on \textit{all} scenarios, vision-based models performed only slightly better when they were trained with the \textit{only} protocol (b=0.21, t=4.4, p<$10^{-4}$), and not significantly worse when trained with the \textit{all-but} protocol (b=0.009, t=1.9, p=0.057, Fig. \ref{fig:model_human_comparision}E).  
Differences between protocols were larger for particle-based models, but nonetheless small relative to overall performance levels.
These results strongly suggest that performance assessments are robust to the specific choices of scenarios we made. 
This makes sense because the diverse physical phenomena in our everyday environment result from a smaller set of underlying laws.  Our results thus quantitatively support the qualitative picture in which an intuitive, approximate understanding of those laws gives rise to humans' outstanding ability to predict and generalize to previously unseen physical phenomena from an early age \citep{spelke1990principles, carey2001infants, baillargeon2011infants, riochet2018intphys}.
However, we do find that models trained on any single scenario do not generalize well to most other scenarios (Fig. \ref{fig:generalization}), suggesting that having substantial diversity of observations is critical for learning general physical forward predictors. 
It will be important, then, to develop additional testing scenarios that incorporate physical phenomena \textit{not} covered here, such as ``squishy'' and fluid materials, the dynamics of jointed multi-part objects, and much larger ranges of mass, friction, density, and other physical parameters.
We thus hope that our benchmark can be used to drive the development of algorithms with a more general, human-like ability to predict how key events will unfold and to anticipate the physical consequences of their own actions in the real world.

\vfill

%% file: Sections/acknowledgments.tex
\section*{Acknowledgments}
D.M.B. is supported by a Wu Tsai Interdisciplinary Scholarship and is a Biogen Fellow of the Life Sciences Research Foundation.
C.H. is supported by a Department of Defense National Defense Science and Engineering Graduate Fellowship.
H.F.T., K.A.S, R.T.P., N.K., and J.B.T are supported by National Science Foundation Science Technology Center Award CCF-1231216 and Office of Naval Research Multidisciplinary University Research Initiative (ONR MURI) N00014-13-1-0333; 
K.A.S. and J.B.T. are supported by research grants from ONR, Honda, and Mitsubishi Electric.
D.L.K.Y is supported by the McDonnell Foundation (Understanding Human Cognition Award Grant No. 220020469), the Simons Foundation (Collaboration on the Global Brain Grant No. 543061), the Sloan Foundation (Fellowship FG-2018-10963), the National Science Foundation (RI 1703161 and CAREER Award 1844724), and hardware donations from the NVIDIA Corporation.
K.A.S., J.B.T., and D.L.K.Y. are supported by the DARPA Machine Common Sense program.
J.E.F. is supported by NSF CAREER Award 2047191 and the ONR Science of Autonomy Program. 
This work was funded in part by the HAI-Google Cloud Credits Grant Program and the IBM-Watson AI Lab.
We thank Seth Alter and Jeremy Schwartz for their help on working with the ThreeDWorld simulator.

%% file: Sections/broaderimpact.tex
\section*{Broader Impact}
There are few aspects of everyday life that are not informed by our intuitive physical understanding of the world: 
moving and doing tasks around the home, operating motor vehicles, and keeping one's body out of harm's way are just a few of the broad behavioral categories that involve making predictions of how objects in the world will behave and respond to our actions.
Although there may be ways for algorithms to safely and effectively perform specific tasks without general, human-like understanding of the physical world, this remains a wide open question in many of the areas where AI is rapidly being deployed: self-driving vehicles, robotics, and other systems that involve a ``perceive-predict-act'' feedback loop.
As such, we think the \textbf{Physion} benchmark is an important step toward actually measuring whether a given algorithm \textit{does} perceive visual scenes and make physical predictions the way people do.
If it turns out that this is critical for achieving safe, high performance in some real-world domain, our benchmark (or its successors) could be used to screen for algorithms more likely to behave like people and to diagnose failures, e.g. by breaking them down into problems making predictions about particular physical phenomena.
Moreover our results, though representing only an initial survey of existing algorithms, \textit{do} suggest that models with more explicit physical representations of the world, including the grouping of scene elements into objects, are better equipped to make accurate predictions;
they therefore begin to address longstanding questions in AI about whether some sort of ``symbolic'' representation, inspired by cognitive science, is necessary for an algorithm to accurately predict and generalize to new situations.
Though such representations have fallen out of favor in large-scale visual categorization tasks, the fact that they outperform their less or non-symbolic counterparts on the \textbf{Physion} tasks raises the intriguing possibility that two broad types of understanding, ``semantic'' and ``physical'', may benefit from different algorithm architectures and learning principles.
If this is the case, we should reevaluate popular claims that symbolic representations and ``interpretable'' algorithms are red herrings for making progress in AI.

%% file: Sections/supplementary.tex
\section{Supplemental Material}

\setcounter{figure}{0}

\makeatletter 
\renewcommand{\thefigure}{S\@arabic\c@figure}
\renewcommand{\thetable}{S\arabic{table}}
\makeatother

\subsection{Adversarial Stimuli and Model-Human Disagreement}

Here we show the distribution of human accuracy on several of the scenarios, which reveals that people are \textit{significantly below chance} on some of the stimuli. Upon investigation, many of these appear to have severe occlusion or are just on the verge of having the opposite trial outcome: a slight change to the initial physical configuration would lead to agent-patient (non)contact. Because \textbf{DPI} is not a vision model, it is insensitive to occlusion; and because it receives ground truth, high-resolution object positions and trajectories as inputs and supervision, it may be less susceptible to the ``observation noise'' that makes certain stimuli ``adversarial'' to humans. For these reasons, there may be an upper bound to how well particle-based models like \textbf{DPI} can match human responses. In addition, \textbf{DPI} and the other particle-based models are deterministic and always make binary predictions; this also limits how well they can match average human decisions, which are typically not 0 or 1. A model with probabilistic learned dynamics or decisions might thus, by averaging over samples, make decisions more like the average person \citep{battaglia2013simulation}.

We have attached 10 randomly sampled stimuli from each scenario at the end of the Supplement. 

\begin{figure}
    \centering
    \includegraphics[width=0.5\textwidth]{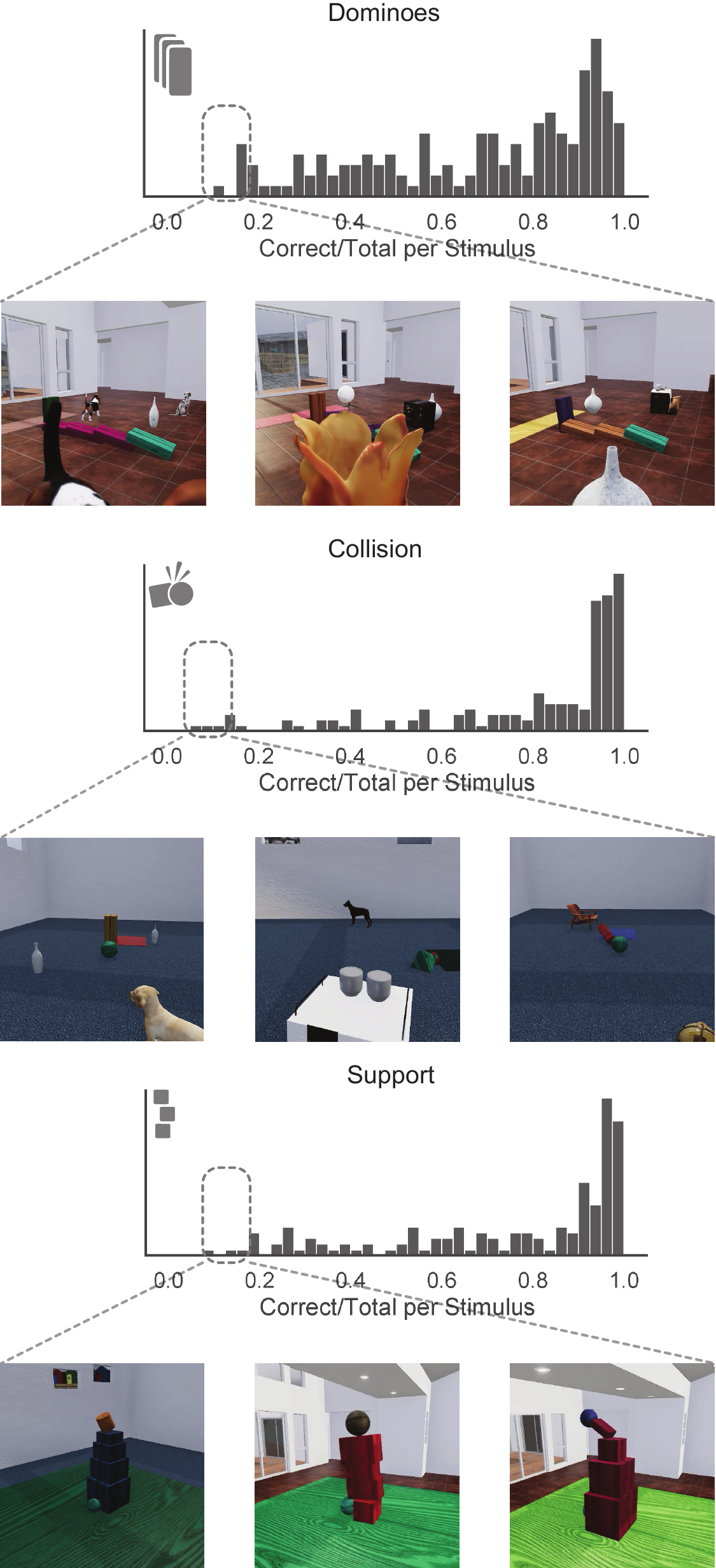}
    \caption{Examples of stimuli on which people performed significantly below chance. The top panel for each scenario shows the per-trial distribution of average human accuracy; sampling from the low end of this distribution gives the examples that are ``adversarial'' for physical prediction. In most cases, these trials are either impossible to get right on average because of occlusion or they are very close to having a different trial outcome: if the initial physical configuration had been just slightly different, the outcome would be the opposite.}
    \label{fig:adversarial}
    \vspace{-0.3cm}
\end{figure}

\begin{figure}
    \centering
    \includegraphics[width=0.99\textwidth]{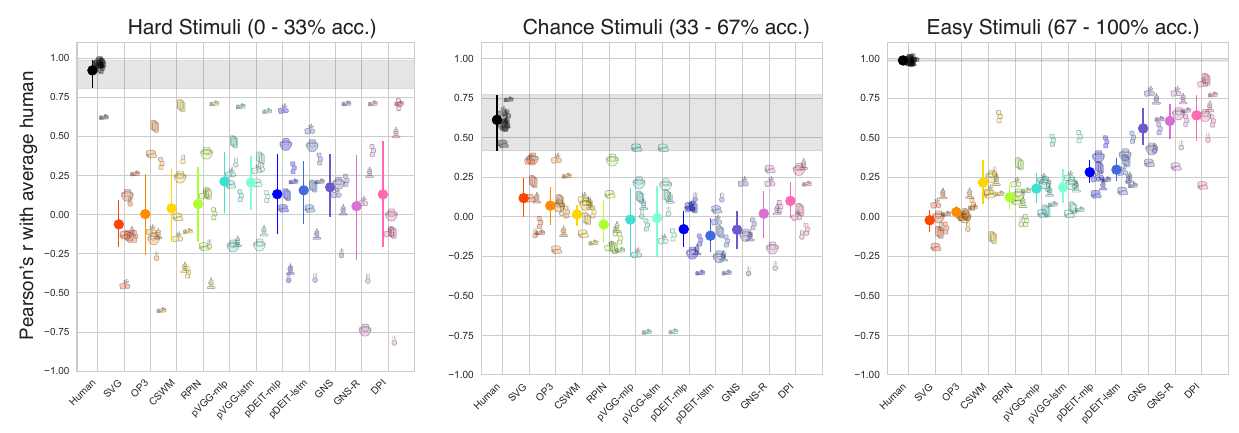}
    \caption{Pearson correlation between model or human responses and the average human response on stimulus subsets defined by average human accuracy. Hard (0 - 33\% accuracy), ``Chance'' (33 - 67\% accuracy), and Easy (67 - 100\% accuracy) stimuli represent 10\%, 22\% and 68\% of the total testing stimuli across all eight scenarios. Gray bars are the ``human zones,'' defined as the 2.5th - 97.5th percentiles of the distribution the correlation between randomly split halves of the human participant pool. Error bars are the 2.5th - 97.5th percentiles of the bootstrapped across-scenario means.}
    \label{fig:hardchanceeasy}
    \vspace{-0.3cm}
\end{figure}

\begin{figure}
    \centering
    \includegraphics[width=0.6\textwidth]{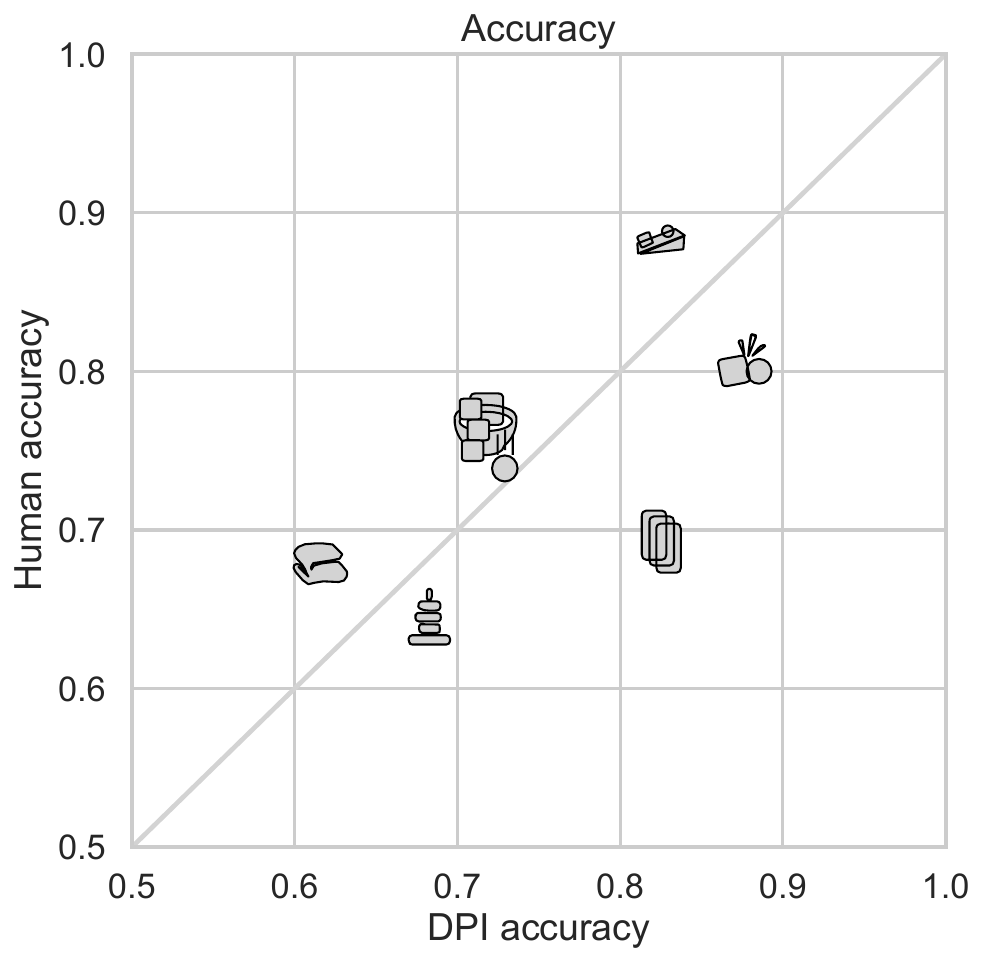}
    \caption{Human and \textbf{DPI} average accuracy across testing stimuli for each scenario. Scenarios below the diagonal indicate super-human performance, but the \textbf{DPI} model is fed ground truth physical inputs and so does not have to contend with occlusion or other limits of visual observation as humans do.}
    \label{fig:dpi_human}
    \vspace{-0.3cm}
\end{figure}

\begin{figure}
    \centering
    \includegraphics[width=\textwidth]{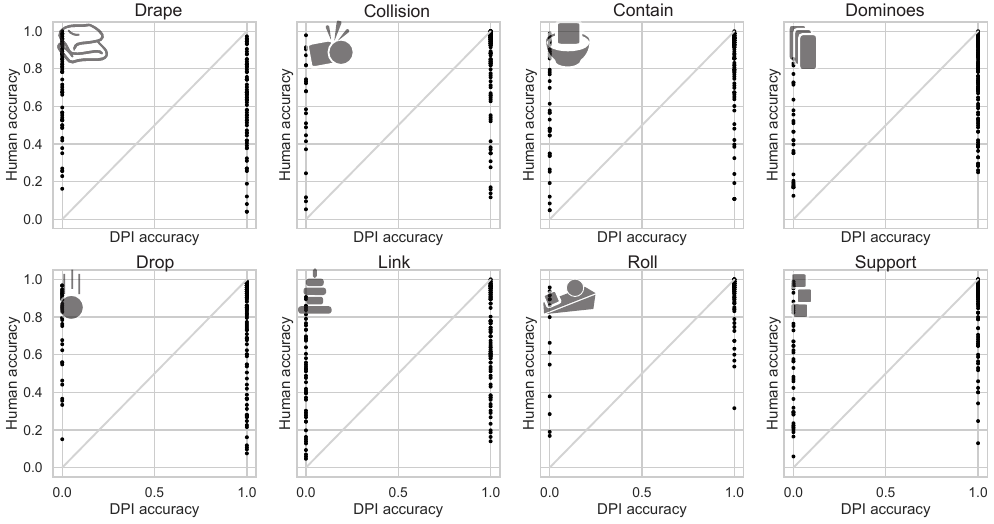}
    \caption{Human accuracy \textit{versus} \textbf{DPI} accuracy per stimulus for each scenario. Each dot is one testing stimulus. Note that \textbf{DPI} makes predictions with the \textit{observed+simulated} readout protocol only, and does so without a context adaptor: there is a fixed distance threshold that determines whether particles from the agent and patient object are in contact at the end of \textbf{DPI}'s learned simulation. As such, this model makes binary predictions, limiting how well correlated its outputs can be with the ``average human'' (real-valued ``average predictions.'') This hints that adding a probabilistic component to \textbf{DPI} and/or non-binarized readout model might lead to a better human-model match.}
    \label{fig:dpi_per_stim}
    \vspace{-0.3cm}
\end{figure}

\subsection{Across Scenario Generalization}

\begin{figure}
    \centering
    \includegraphics[width=\textwidth]{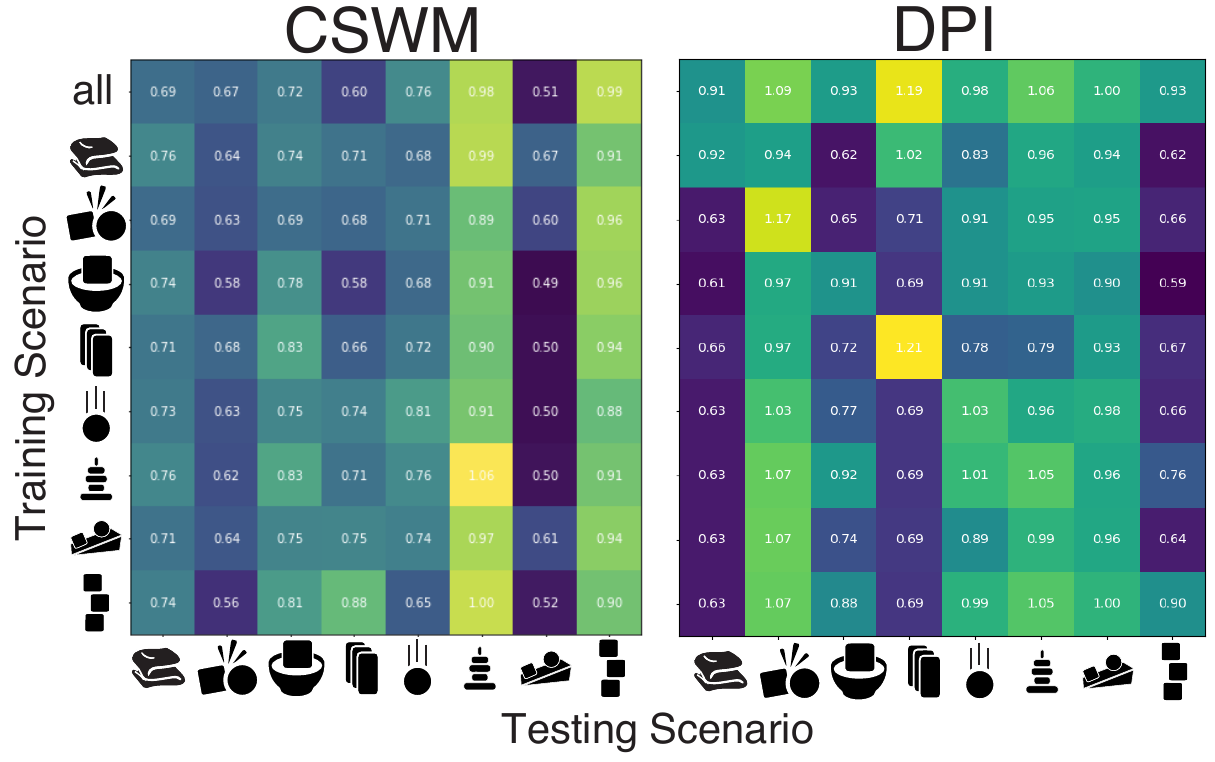}
    \caption{Performance on each scenario's testing set when \textbf{CSWM} (left) or \textbf{DPI} (right) were trained on each of the scenarios or all of them combined. Color and value for each cell indicate performance relative to the average human on that scenario. For \textbf{DPI}, training on any single scenario gave near-human performance on \textbf{Collide} and \textbf{Roll}, and training on most single scenarios gave near-human performance on \textbf{Drop} and \textbf{Link}. However, no single training scenario was suitable for generalization to all others, compared to training on all the scenarios. \textbf{Drape} and \textbf{Support} training appeared to yield the best generalization, perhaps because the ground truth dynamics of these scenarios include many soft and rigid object-object interactions at a wide range of velocities.}
    \label{fig:generalization}
    \vspace{-0.3cm}
\end{figure}

In addition to the \textit{all}, \textit{all-but}, and \textit{only} training protocols, we tested the ``best'' TDW-trained vision model (\textbf{CSWM}) and particle model (\textbf{DPI}) for their ability to generalize from any single scenario to any other scenario (Fig. \ref{fig:generalization}). 
Generalization was fairly homogeneous across training sets for \textbf{CSWM}, but this may merely reflect poor overall performance.
For \textbf{DPI}, clearer patterns emerged: some scenarios were hard to do well on unless they were in the training set (\textbf{Drape}, \textbf{Dominoes}, \textbf{Support}) whereas training on almost \textit{any} scenario was sufficient to give good performance on \textbf{Drop}, \textbf{Link}, \textbf{Roll}, and especially \textbf{Collide}.
However, no single scenario made for as strong a training set as combining \textbf{all} of them; \textbf{Drape} and \textbf{Support} came the closest, perhaps because they include many object-object interactions in every trial.
Overall these data suggest that the eight scenarios cover many distinct physical phenomena, such that experience with any one is insufficient to learn a good prediction model;
on the other hand, some phenomena (like object-object contact) may be so ubiquitous that the scenarios with more of them are simply better for efficiently learning about physics in general.
The diversity of train-test ``fingerprints'' for even the most human-like model, combined with the fact that training on \textbf{all} scenarios gives the best across-the-board performance, implies that our chief desideratum for the \textbf{Physion} benchmark was a crucial choice: developing algorithms on only one or a few physical scenarios would not have produced nearly as general prediction models.

\subsection{Model Performance Per Scenario}

Table~\ref{tab:mod_scenario} shows model accuracies for every model in each of the eight scenarios, as compared to human performance. There is heterogeneity in performance across the scenarios, with some scenarios (e.g., \textbf{Roll}) that people find easy but for which no model approaches human performance, and other scenarios (e.g., \textbf{Link}) that people find difficult, but where model accuracy approaches or exceeds humans.

\begin{table}[ht]
\caption{Model and human accuracy for each of the eight different scenarios. Numbers indicate mean accuracy with bootstrapped 95\% confidence intervals. Italicized values represent instances where the models perform reliably worse than people; bold values represent instances where the models perform reliably better.}
\label{tab:mod_scenario}
\small
\centering
\begin{tabular}{lllll}
  \hline
Model & Dominoes & Support & Collide & Contain \\ 
  \hline
Human & 0.693 & 0.763 & 0.809 & 0.767 \\ 
  SVG & \textit{0.538 [0.512, 0.565]} & \textit{0.596 [0.574, 0.619]} & \textit{0.597 [0.58, 0.612]} & \textit{0.56 [0.545, 0.576]} \\ 
  OP3 & \textit{0.47 [0.457, 0.485]} & \textit{0.516 [0.504, 0.529]} & \textit{0.511 [0.501, 0.522]} & \textit{0.499 [0.488, 0.509]} \\ 
  CSWM & \textit{0.471 [0.432, 0.519]} & \textit{0.691 [0.636, 0.748]} & \textit{0.552 [0.528, 0.577]} & \textit{0.557 [0.523, 0.593]} \\ 
  RPIN & \textit{0.625 [0.61, 0.641]} & \textit{0.62 [0.591, 0.651]} & \textit{0.645 [0.617, 0.674]} & \textit{0.601 [0.576, 0.627]} \\ 
  pVGG-mlp & 0.601 [0.505, 0.7] & \textit{0.669 [0.631, 0.708]} & \textit{0.651 [0.608, 0.7]} & \textit{0.638 [0.595, 0.684]} \\ 
  pVGG-lstm & 0.603 [0.513, 0.7] & \textit{0.675 [0.641, 0.711]} & \textit{0.651 [0.606, 0.699]} & \textit{0.643 [0.599, 0.693]} \\ 
  pDEIT-mlp & 0.664 [0.572, 0.757] & \textit{0.686 [0.636, 0.736]} & \textit{0.677 [0.633, 0.721]} & \textit{0.664 [0.645, 0.684]} \\ 
  pDEIT-lstm & 0.664 [0.572, 0.767] & \textit{0.687 [0.637, 0.739]} & \textit{0.681 [0.637, 0.727]} & \textit{0.669 [0.654, 0.684]} \\ 
  GNS & 0.604 [0.477, 0.859] & \textit{0.695 [0.674, 0.711]} & 0.85 [0.804, 0.912] & \textit{0.652 [0.62, 0.702]} \\ 
  GNS-R & 0.591 [0.477, 0.819] & \textit{0.686 [0.619, 0.732]} & 0.842 [0.808, 0.908] & 0.683 [0.512, 0.776] \\ 
  DPI & 0.715 [0.477, 0.841] & \textit{0.626 [0.477, 0.711]} & 0.85 [0.725, 0.946] & \textit{0.711 [0.698, 0.717]} \\ 
   \hline
   Model & Drop & Link & Roll & Drape \\ 
  \hline
Human & 0.744 & 0.643 & 0.883 & 0.678 \\ 
  SVG & \textit{0.533 [0.52, 0.548]} & \textit{0.544 [0.53, 0.558]} & \textit{0.561 [0.545, 0.577]} & \textit{0.545 [0.532, 0.559]} \\ 
  OP3 & \textit{0.526 [0.512, 0.541]} & \textit{0.545 [0.54, 0.551]} & \textit{0.544 [0.529, 0.559]} & \textit{0.548 [0.523, 0.57]} \\ 
  CSWM & \textit{0.577 [0.542, 0.613]} & 0.627 [0.603, 0.649] & \textit{0.609 [0.587, 0.632]} & \textit{0.55 [0.496, 0.605]} \\ 
  RPIN & \textit{0.551 [0.538, 0.564]} & \textit{0.597 [0.58, 0.614]} & \textit{0.622 [0.604, 0.638]} & \textit{0.596 [0.585, 0.608]} \\ 
  pVGG-mlp & \textit{0.606 [0.577, 0.639]} & 0.614 [0.581, 0.649] & \textit{0.573 [0.548, 0.6]} & \textit{0.6 [0.572, 0.63]} \\ 
  pVGG-lstm & \textit{0.603 [0.572, 0.638]} & 0.618 [0.583, 0.657] & \textit{0.573 [0.546, 0.602]} & \textit{0.599 [0.571, 0.629]} \\ 
  pDEIT-mlp & \textit{0.619 [0.589, 0.651]} & \textit{0.59 [0.546, 0.633]} & \textit{0.62 [0.601, 0.642]} & \textit{0.608 [0.586, 0.631]} \\ 
  pDEIT-lstm & \textit{0.614 [0.582, 0.65]} & \textit{0.592 [0.55, 0.639]} & \textit{0.616 [0.597, 0.638]} & \textit{0.608 [0.586, 0.633]} \\ 
  GNS & \textit{0.708 [0.69, 0.74]} & \textbf{0.73 [0.707, 0.756]} & \textit{0.735 [0.718, 0.752]} & 0.653 [0.598, 0.714] \\ 
  GNS-R & \textit{0.712 [0.7, 0.735]} & \textbf{0.725 [0.717, 0.737]} & \textit{0.792 [0.752, 0.872]} & 0.653 [0.598, 0.714] \\ 
  DPI & 0.755 [0.73, 0.77] & 0.657 [0.615, 0.683] & \textit{0.789 [0.769, 0.821]} & \textit{0.556 [0.432, 0.623]} \\ 
   \hline
\end{tabular}
\end{table}

\subsection{Model Details}

Here we describe the four classes of model we test and provide implementation and training details for the representatives we selected. If not stated otherwise, models' visual encoder and/or dynamics predictor architectures were unchanged from their published implementations.

\paragraph{i. Unsupervised visual dynamics models.}
These are models explicitly designed to learn dynamical, predictive representations of the visual world without ground truth supervision on physical scene variables.
We further divide them into two types: models with \textit{image-like latent representations} and models with \textit{object-like latent representations}.
Our representative from the first type, SVG \citep{denton2018svg}, uses a convolutional encoder $\mathcal{E}$ to predict a latent hidden state $\mathbf{p}$, then uses (a) an LSTM-based dynamics model based on the hidden state and a randomly sampled latent from a learned prior distribution to predict a future hidden state $\mathbf{q}$ and (b) a hidden-state-to-image decoder to predict a future frame of the input movie, $\hat{X}_{t_{pred}}$. The model is trained by optimizing the variational lower bound. 
\textbf{SVG} is trained on movies from the benchmark; testing this model therefore tests whether physical understanding can emerge from a convolutional future prediction architecture, without imposing further constraints on the structure of the learned latent representation of scenes or dynamics.

Our representatives with object-like latent representations are \textbf{CSWM} and \textbf{OP3}. 
These models were designed under the hypothesis that physical understanding requires a decomposition of scenes into objects.
We call these representations ``object-like'' rather than ``object-centric'' because the latent variables are not explicitly constrained to represent physical objects; they are merely encouraged to do so through the models' inductive biases and unsupervised learning signals.
Specifically, both \textbf{CSWM} and and \textbf{OP3} use convolutional encoders $\mathcal{E}$ to predict \textit{K-factor latent representations},
\begin{equation}
    \mathbf{p} := \mathbf{o}_1 \oplus \mathbf{o}_2 \oplus ... \oplus \mathbf{o}_{K},
\end{equation}
where each inferred \textit{object vector} $\mathbf{o}_k \in \mathbb{R}^{t_{vis} \times P}$ is meant to encode information about one and only one object in the observed scene.
The dynamics models for \textbf{CSWM} and \textbf{OP3} are \textit{recurrent graph neural networks} that pass messages between the object vectors at each iteration of future prediction to produce a new set of predicted object vectors,
\begin{equation}
    \mathcal{D}_{\theta_d} \equiv 
    \mathcal{G}^{(t_{pred})}_{\theta_d}: \mathbf{p}[t_{vis},:,:] \mapsto
    \mathbf{\hat{o}}_1 \oplus \mathbf{\hat{o}}_2 \oplus ... \oplus \mathbf{\hat{o}}_{K}
    \equiv \mathbf{q},
\end{equation}
where the graph neural network $\mathcal{G}$ is iterated $t_{pred}$ times to produce as many estimates of the future object states.
\textbf{OP3} learns the parameters $\theta_e \cup \theta_d$ by applying a \textit{deconvolutional decoder} to render the future object states into a predicted future movie frame, which is used to compute an L2 loss with the actual future frame.
\textbf{CSWM} instead learns these parameters with a contrastive hinge loss directly on the predicted object-like latent state $q$; see \citep{kipf2019contrastive} for details.
Thus, these models test whether physical understanding can emerge by predicting scene dynamics through a representation architecture with discrete latent factors, which \textit{could} represent properties of individual objects in the scene but are not explicitly constrained to do so.

\paragraph{ii. Supervised visual-physical dynamics models.}
We next asked whether vision models with an \textit{explicit object-centric representation}, rather than merely an ``object-like'' representation, would be better suited for physical understanding.
Our representative model from this class was \textbf{RPIN} \citep{qi2020learning}. Region Proposal Interaction Networks (\textbf{RPIN}) take a short sequence of \emph{N} frames as inputs and output the future 2D object positions on the image. The sequence of frames is passed through an encoder network based on a R-CNN like object detection architecture \citep{girshick2015fast} which uses RoIPooling to extract object-centric features from the images. A sequence of \emph{k} object features is then forwarded to an interaction network \citep{battaglia2016interaction} to resolve object and environment interactions and predict the future object features at the next time step. The future object features are then decoded to the individual 2D object locations on the image. To be able to estimate velocity and acceleration, we use 4 input images to the interaction network based physics predictor. 
In contrast to the unsupervised models in section \textbf{i}, supervision in the form of human annotated bounding boxes is required to train the RoIPooling based encoder and object location decoder.
Thus this model is much more constrained than the models in \textbf{i} to represent scenes as a set of discrete objects whose positions change smoothly over time.
Although it is not a realistic model of how humans \textit{learn} about the physical world without ground truth supervision, success on our benchmark with \textbf{RPIN} where other models failed would strongly suggest that explicit, spatial object-centric representations are useful for intuitive physical understanding of scenes.

\paragraph{iii. Pretrained visual encoders.}
These visual encoders are optimized to perform a challenging vision task, such as object classification.
Although these tasks are not directly related to intuitive physics, it is possible that machine learning models only solve them by learning some partial, implicit representation of the physical world.
We tested two models, the standard Convolutional Neural Network VGG-19 and a newer model with a Transformer-based architecture, DeIT, both trained on the supervised ImageNet task.
In our decomposition, these models consist only of pretrained encoders $\mathcal{E}_{\theta_e}$ that take $t_{vis}$ independent movie frames as input and produce an output feature vector
\begin{equation}
    \mathbf{p}_{1:t_{vis}} := \mathbf{v}_1 \oplus \mathbf{v}_2 \oplus ... \oplus \mathbf{v}_{t_{vis}},
\end{equation}
where $\mathbf{v}_t$ is the vector of activations from the penultimate layer of the encoder on frame $t$.
These were not designed to do explicit physical simulation and thus have no dynamics model $\mathcal{D}_{\theta_d}$.
We therefore provide them with simple dynamics models that can be ``rolled out'' a variable number of time steps,
\begin{equation}
    \mathcal{D}_{\theta_{d}}: \mathbf{p}_{1:t} \mapsto \mathbf{w}_{t+1},
\end{equation}
where $\mathcal{D}_{\theta_{d}}$ is a MLP for \textbf{pVGG/pDeIT-mlp} and a LSTM for \textbf{pVGG/pDeIT-lstm}, both with a single hidden layer.
The encoder parameters $\theta_{e}$ are \textit{frozen} and the dynamics model parameters $\theta_d$ are trained with an \textit{unsupervised forward prediction} L2 loss on the unlabeled benchmark training datasets.
Thus, dynamics training and evaluation of these models tests whether their pretrained representations contain latent information useful for physical understanding.

\paragraph{iv. Physical state-computable dynamics models.}
Finally, we consider several models that are not computer vision algorithms at all: 
rather than taking a movie of RGB frames $\{X_{1:t_{vis}}\}$ as input, they take (a subset of) the ground truth simulator state, $\{S_{1:t_{vis}}\}$ and make predictions about how it will evolve over time, supervised on the ground truth future states.
The point of testing these non-visual models is to isolate two distinct challenges in physical understanding: (1) representing some of the physical structure of the world from visual observation (captured by encoding models $\mathcal{E}$) and (2) understanding how that structure behaves (captured by dynamics models $\mathcal{D}$).
If models given the ground truth physical state -- i.e., models that did not have to solve challenge (1) -- matched human performance on our benchmark, we would conclude that the major objective for physical understanding research should be addressing the visual representation problem.
On the other hand, if these pure dynamics models still did not match human performance, we would conclude that problem (2) remains open and would benefit from alternative proposals and tests of how people represent and use intuitive physical knowledge about scenes.
Thus, comparing these physically explicit, supervised models with those in \textbf{i - iii} illustrates how to use our benchmark to diagnose key issues in machine physical understanding.

We consider two graph neural network architectures of this kind, DPI-Net (\textbf{DPI}) \citep{li2018learning} and \textbf{GNS} \cite{sanchez2018graph}. 
Both models operate on a \textit{particle graph representation} of scenes, which for our dataset is constructed by taking the ground truth collider meshes of each object, converting each mesh vertex into a leaf-level graph node (i.e., particle), and connecting these particles \textit{via} edges that represent physical connections. For GNS, edges are dynamically constructed by adding edges between 2 particles that have distance smaller than a threshold, $\delta.$ $\delta$ is set to 0.08 for all model variations. For \textbf{DPI}, aside from connecting particles with small enough distance, particles belonging to the
same object is connected with an object-level root node. The root node can help propagate effect from far away particles within the same object. 
The DPI-Net run in our experiments differs from the original implementations in two ways: (1) we use relative particle positions, as opposed to absolute particle positions, to improve model generalization, as suggested in \textbf{GNS} \cite{sanchez2018graph}. (2) The original DPI-Net does not include any leaf-leaf edges between particles within an object. We find out excluding such edges leads to bad performance on objects with a large number of particles. To handle objects with diverse number of particles in our dataset, we include these within object edges that indicates close-by particles.

Both \textbf{DPI} and \textbf{GNS} explicitly represent each particle's 3D position and instantaneous velocity at each movie frame and make predictions about these node attributes' future values using a rolled out graph neural network, which at each iteration passes learned messages between particles that depend on their attributes and the presence or absence of an edge between them.
The key difference between the two models is that DPI-Nets operate on graphs with 2-level hierarchy (, i.e., graph with leaf-level nodes and root-level nodes) while \textbf{GNS} operates on flat graphs with no hierarchy. We observe that \textbf{GNS} can make good prediction even without explicitly modeling the hierarchy explicitly, yet the objects tend to deform during long-term forward unrolling, due to error accumulation over time. These deformed objects can trigger the models to generate unreasonable
predictions such as having all the particles scattering and floating in the free space. To solve the problem, we further include a model variation called GNS-RANSAC (\textbf{GNS-R}) that tries to enforce rigid objects to be rigid over time. During model forward unrolling for \textbf{GNS}, we run RANSAC \cite{ransac} on top of each object to compute the 6-Dof rotation and translation matrix for the object and use the matrix to compute the updated positions for the object's particles.

\subsection{Experimental Details} \label{A2:computational_details}

Experiments were run on Google Cloud Platform (GCP) across 80 GPUs (NVIDIA T4s \& V100s) for two days. DPI-Nets and GNS are trained for 1.5M~2M iterations till converge using Adam optimizer with initial learning rate 1e-4. Experiments take around 2-5 days to train.

\subsection{Links to access the dataset and its metadata.}

\subsection{Long-term preservation plan}

\subsection{License Information} \label{license}
All products created as part of this project is shared under the MIT license (including code \emph{and} data), and this license has been uploaded to the Github repo where our code is stored and our data is referenced. 

We used a number of third-party software packages, each of which typically has its own licensing provisions.  Table \ref{table:licenses} contains a list of these licenses for many of the packages used.
\begin{center}
    \begin{table}
        \caption{Table of open-source code used.}
        \begin{tabular}{||c c c||} 
            \hline
            Name & URL & License \\ [0.5ex] 
            \hline\hline
            SVG\citep{denton2018svg} & \url{https://github.com/edenton/svg} & N/A\\
            C-SWM\citep{kipf2019contrastive} & \url{https://github.com/tkipf/c-swm} & MIT License\\
            OP3\citep{veerapaneni2020entity} & \url{https://github.com/jcoreyes/OP3} & MIT License\\
            RPIN \citep{qi2020learning} & \url{https://github.com/HaozhiQi/RPIN} & N/A\\
            DeIT \citep{touvron2020deit} & \url{https://github.com/facebookresearch/deit} & Apache License 2.0\\
            VGG \citep{simonyan2014vgg, pytorch} & \url{https://github.com/pytorch/vision} & BSD 3-Clause License \\
            DPI-Net \citep{li2018learning} & \url{https://github.com/YunzhuLi/DPI-Net} & N/A \\
            TDW \citep{gan2020threedworld} & \url{https://github.com/threedworld-mit/tdw} & BSD 2-Clause License\\
            \hline
        \end{tabular}
        \label{table:licenses}
    \end{table}
\end{center}

\subsection{Datasheets for dataset}
Here are our responses in reference to the Datasheets for Datasets~\cite{gebru2018datasheets} standards.

\subsubsection{Motivation}
\begin{itemize}
    \item \textbf{For what purpose was the dataset created?} To measure adult human short-term physical future prediction abilities and compare these to predictions made by AI models. 
    \item \textbf{Who created the dataset and on
behalf of which entity?} The authors listed on this paper, including researchers from Stanford, UCSD, and MIT.
    \item \textbf{Who funded the creation of the dataset?} The various granting agencies supporting the above-named researchers, including both grants to the PIs as well as individual fellowships for graduate students and postdoctoral fellows involved with the project.  A partial list of funders includes the NSF, NIH, DARPA, and the McDonnell Foundation.
\end{itemize}

\subsubsection{Composition}
\begin{itemize}
    \item \textbf{What do the instances that comprise the dataset represent?} Each instance is a video of a simulated physical scene (e.g. a tower of blocks as it either collapses or remains steady), together with some metadata about that video, including map-structured metadata with depth maps, normal maps, object instance maps, \&c, and information about object-object collisions at each timepoint. 
    \item \textbf{How many instances are there in total?} The dynamics prediction model training dataset consists of 2000 examples for each of the 8 scenarios.  The OCP readout fitting dataset consists of 1000 examples per each of the 8 scenarios. The test dataset (on which human responses were obtained) consists of 150 examples per scenario. 
    \item \textbf{Does the dataset contain all possible instances or is it a sample of instances from a larger set?} Data is generated by a simulator; in a sense, the set of datapoints we created is an infinitesimally small subset of data that \emph{could} have been generated.  However, we are all here releasing all the examples we did actually generate.
    \item \textbf{What data does each instance consist of?} It consists of a video depicting a physical situation (e.g a tower of blocks falling over), together with simulator-generated metadata about the situation.
    \item \textbf{Is there a label or target associated with each instance?} For the training dataset, there are no labels.  For both the OCP readout fitting dataset and the human testing dataset, there are binary labels describing whether the red object collided with the yellow zone during the duration of the trajectory. 
    \item \textbf{Is any information missing from individual instances?} No. 
    \item \textbf{Are relationships between individual instances made explicit?} Yes. All data is provided in a simple data structure that indicates which instances of data are connected with which instances of metadata. 
    \item \textbf{Are there recommended data splits?} Yes, for each of the scenarios in the datasets, there are three splits:  (a) a large training split for training physical prediction models from scratch; (b) a smaller readout-training set that is to be used for training the yes/no binary readout training as described in the paper, and (c) the test dataset on which human responses were obtained.  
    \item \textbf{Are there any errors, sources of noise, or redundancies in the dataset?} Probably, but we don't know if any at the moment. As these are discovered, they will be fixed and versioned. 
    \item \textbf{Is the dataset self-contained, or does it link to or otherwise rely on external resources?} It is self-contained.
    \item \textbf{Does the dataset contain data that might be considered confidential?} No. 
    \item \textbf{Does the dataset contain data that, if viewed directly, might be offensive, insulting, threatening, or might otherwise cause anxiety?} No. 
    \item \textbf{Does the dataset relate to people?} No. 
\end{itemize}

\subsubsection{Collection Process}
\begin{itemize}
    \item \textbf{How was the data associated with each instance acquired? What mechanisms or procedures were used to collect the data? How was it verified?}  Videos (for training, readout fitting, and human testing) were generated using the TDW simulation environment.  Online crowdsourcing was used to obtain human judgements for each testing video.  During the creation of the simulated videos, the researchers looked at the generated videos by eye to verify if the scenarios were correct (e.g. actually depicted the situations desired by our experimental design). Prior to running the actual data collection procedure for humans, we verified that the experimental websites were correct by having several of the researchers complete the experiment themselves. 
    \item \textbf{Who was involved in the data collection process and how were they compensated?} PIs, students, and postdocs generated simulator-generated videos.  Human responses were obtained via the Profilic  platform, and subjects where compensated \$4 for participation.  
    \item \textbf{Over what timeframe was the data collected?} All simulator-generated scenarios were created during early May 2021.  All human data was collected during approximately one week in May 2021. 
    \item \textbf{Were any ethical review processes conducted?} All human data collection was approved by Stanford and UCSD IRBs. 
    \item \textbf{Does the dataset relate to people?} No.
\end{itemize}

\subsubsection{Preprocessing, clearning and labelling.}

\begin{itemize}
    \item \textbf{Was any preprocessing/cleaning/labeling of the data done?} No. All our input data was  simulator-generated (so we knew the labels exactly and could avoid any cleaning procedures).  The comparison between model and human responses is made directly on the raw collected human judgements with no further preprocessing.
\end{itemize}

\subsubsection{Uses.}
\begin{itemize}
    \item \textbf{Has the dataset been used for any tasks already?} Yes, the participants in the human experiments used the data for the single purposes for which it was designed: obtaining detailed characterization of human judgements about short-term physical prediction in simple scenes. 
    \item \textbf{Is there a repository that links to any or all papers or systems that use the dataset?}. No other papers use the dataset yet. 
    \item \textbf{What (other) tasks could the dataset be used for?} None. 
    \item \textbf{Is there anything about the composition of the dataset or the way it was collected and preprocessed/cleaned/labeled that might impact future uses?} No. 
    \item \textbf{Are there tasks for which the dataset should not be used?} The dataset can only be used to measure abilities of humans or models to make short-term forward predictions about simple physical scenarios. 
\end{itemize}

\subsubsection{Distribution.}
\begin{itemize}
    \item \textbf{Will the dataset be distributed to third parties outside of the entity (e.g., company, institution, organization) on behalf of which the dataset was created?} Yes it will be completely publicly available via a github repo and the links listed thereupon. 
    \item \textbf{How will the dataset will be distributed?} It will be available on Github (where code for dataset generation will be available, and via links to the raw human data that will be listed on that Github repo, and which will refer to permanent Amazon S3 resources. 
    \item \textbf{When will the dataset be distributed?} Immediately. 
    \item \textbf{Will the dataset be distributed under a copyright or other intellectual property (IP) license, and/or under applicable terms of use (ToU)?} The dataset and associated code will be licensed under the MIT license. 
    \item \textbf{ Have any third parties imposed IP-based or other restrictions on the data associated with the instances?} No. 
    \item \textbf{Do any export controls or other regulatory restrictions apply to the dataset or to individual instances?} No. 
\end{itemize}

\subsubsection{Maintenance}

\begin{itemize}
    \item \textbf{Who is supporting/hosting/maintaining the dataset?} Code for dataset generation will be hosted in GitHub, via a publicly-accessible repo.  The Github account with which this repo is associated is the institutional account for the CogTools lab (at UCSD). 
    \item \textbf{How can the owner/curator/manager of the dataset be contacted?} The corresponding author of the paper can be contacted via email as described in the front page of the paper. 
    \item \textbf{Is there an erratum?} Not yet, but there may be in the future. 
    \item \textbf{Will the dataset be updated (e.g., to correct labeling errors, add new instances, delete instances)?} Yes, we expect the dataset to be expanded over the next few months or so.  Errors will be corrected as they are discovered on an ongoing basis.  Updates will be communicated to users via notes on the commits to the Github repo. 
    \item \textbf{Will older versions of the dataset continue to be supported/hosted/maintained?} If newer versions of the dataset are created, these will only be in additional to the existing data. Old versions will be maintained indefinitely. 
    \item \textbf{If others want to extend/augment/build on/contribute to the dataset, is there a mechanism for them to do so?} No. Making contributions to this dataset requires very substantial expertise in psychophysical experimental design, and we do not contemplate allowing third parties to (e.g.) add new examples of physical scenarios.  Of course, the code for generating the data and for setting up crowd-sourced psychophysical collection is completely open source, so others could easily fork our repos and make their own versions of such benchmarks of they choose.
\end{itemize}

\subsection{Structured metadata}
We have not created structured metadata for our project in a format like that in schema.org or DCAT as yet, because we expect that through the review feedback process, the exact structure of what metadata we should provide may change a bit. We'd be happy to do this once review is complete. In the meantime, all of our data is available through our github repo, which provides a certain level of metadata bout the project that we think is appropriate for the review process. 

\subsection{Dataset identifier}
Our project provides two types of resources:  a dataset and a set of code for creating / analyzing the data.   At the moment, we provide access to the code via the GitHub repo, and to the data via Amazon S3 links that are visible via the GitHub repo.  We have not yet pushed out data into a standard data repository or created a DOI for it.   This is because we expect the specifics of how the data is made available to develop a bit via the paper review process.  Once this is complete, we will push the data into a standardized data repository and generate a DOI for it.

\section{Human experimental study preregistration}

This analysis plan was prepared according to the suggested template for experimental study preregistration documents from the Open Science Framework.

\subsection{Study information}\label{study-information}

\textbf{Title}: Human physics benchmarking

\subsubsection{Research questions}\label{research-questions}

Predicting the future outcome of physical scenarios is a paradigm case
of using models to represent and reason about the world. Intuitive
physics is central to intelligent behavior in physical environments. In
this study, we aim to identify features of physical scenes that make
correct human physical prediction difficult. Additionally, we aim to
collect data on which scenes are difficult for human participants to
predict correctly in order to compare human participants against a range
of computational models of physical scene prediction.

\subsubsection{Hypotheses}\label{hypotheses}

We predict that scenes which (1) contain more elements, (2) contain
distractor elements and (3) contains occluder elements are harder to
correctly predict for human participants. Additionally (4), we predict
that scenes that lead to more incorrect predictions also tend to have a
longer reaction time (ie. people take longer to come up with an answer
to difficult scenes).

\subsection{Design Plan}\label{design-plan}

\subsubsection{Study design}\label{study-design}

We conducted 8 experiments, each testing physical judgments for different categories of physical scenarios. 

Scenes are generated by sampling values of various physical parameters (e.g., 
number of physical elements, number of occluder objects, positional jitter, etc.) and generating a stimulus set containing \textgreater{}150 example scenes. From this set, 150 will be randomly sampled such that 50\% of the chosen scenes are positive trials (ie. the red target object touches the yellow target
zone) and 50\% are negative trials. Additionally, we attempt to sample
scenes such that the distribution of the other dimensions is roughly
equal if possible. Stimuli will be manually checked to ensure that all
scenes are usable, do not contain off screen elements, exhibits bugs in
the physics engine, contain clipping objects, etc.

\paragraph{Manipulated variables}\label{manipulated-variables}

 As outlined above, participants are not assigned to any conditions. The
manipulations consist of the stimuli with underlying parameters as well
as the sampling of stimuli.

\subsubsection{Study design: evaluation
protocol}\label{study-design-evaluation-protocol}

\textbf{Sequence of events in a session} 1. Consent form and study
information 2. Task explanation 3. Familiarization trials -- 10 shown 1.
First frozen frame shown for 2000ms, with red/yellow segmentation map
indicating agent/patient object flashing at 2Hz 2. Video is played for
1500ms, then hidden 3. Prediction is queried from subject (yes/no) 4.
Full video is shown and feedback is given (correct/incorrect) 5.
Participants can proceed after full video has played 5. Participants are
informed that the main trial starts 6. 100 trials 1. Fixation cross is
shown for random interval between 500ms and 1500ms 2. First frozen frame
shown for 2000ms, with red/yellow segmentation map indicating
agent/patient object flashing at 2Hz 3. Video is played for 1500ms, then
hidden 4. Prediction is queried from subject (yes/no) 7. Demographics \&
Feedback * age * gender * education level * difficulty rating (``How
difficult did you find this task?'', 5 point Likert scale) 8.
Participants are shown their rate of correct guesses 9. End of study

Each stimulus consists of a short video clip of a visual scene
containing various objects physically interacting with each other. Each
of these 150 trials began with a fixation cross, which was shown for a
randomly sampled time between 500ms and 1500ms. To indicate which of the
objects shown is the agent and patient object, participants were then
shown the first frame of the video for 2000ms. During this time, the
agent and patient objects were overlaid in red and yellow respectively.
The overlay flashed on and off with a frequency of 2Hz. After this, the
first 1500ms of the stimulus were played. After 1500ms, the stimulus is
removed and the response buttons are enabled. The experiments moved to
the next phase after the participants made a prediction by selecting
either ``YES'' or ``NO.''

Participants first completed 10 familiarization trials before moving on
to complete 150 test trials. During the familiarization phase, all
participants were presented with the same sequence of stimuli and were
provided with feedback indicating whether their prediction was correct
and were shown the unabridged stimulus including the result of the
trial. During the test phase, participants were presented with the same
set of stimuli in a randomized sequence, and were not provided with
accuracy feedback nor did they observe the subsequent video frames in
the scenario.

\subsubsection{Measured variables}\label{measured-variables}

 We measure: * \texttt{response}: prediction (either yes/no) *
\texttt{rt}: time taken to make prediction

After the trials, participants will be asked to provide: * age * gender
* education level * difficulty rating (``How difficult did you find this
task?'', 5 point Likert scale) * free form feedback on the task

After the end of the study, participants will be told their overall
accuracy and the corresponding percentile compared to other participants
on the study.

\subsection{Sampling Plan}\label{sampling-plan}

\subsubsection{Data collection
procedure}\label{data-collection-procedure}

 Participants will be recruited from Prolific and compensated \$4, which
roughly corresponds to \$12/hr. participants will not be rewarded for
correct responses.

Participants are only allowed to take the task once. However,
participants are able to take a version of the experiment with another
scenario.

\subsubsection{Sampling procedure}\label{sampling-procedure}

 Data collection will be stopped after 100 participants have completed
the experiment.

\subsection{Analysis Plan}\label{analysis-plan}

\subsubsection{Data exclusion criteria}

 Data from an entire experimental session will be excluded if the
responses: * contain a sequence of greater than 12 consecutive ``yes''
or 12 consecutive ``no'' answers (based on simulations run with
p(yes)=0.5) * contain a sequence of at least 24 trials alternating
``yes'' and ``no'' responses * are correct for fewer than 4 out of 10
familiarization trials (i.e., 30\% correct or lower) * the mean accuracy
for that participant is below 3 standard deviations below the median
accuracy across all participants for that scenario * the mean
log-transformed response time for that participant is 3 standard
deviations above the median log-transformed response time across all
participants for that scenario

Excluded sessions will be flagged. Flagged sessions will not be included
in the main analyses. We will also conduct our planned analyses with the
flagged sessions included to investigate the extent to which the
outcomes of the main analyses change when these sessions are included.
Specifically, we will fit a statistical model to all sessions and
estimate the effect of a session being flagged on accuracy.

\subsubsection{Missing data}\label{missing-data}

We will only include sessions that are complete (i.e., response
collected for all trials) in our main analyses.

\subsubsection{Planned analyses}

\paragraph{Human accuracy across participants for each
stimulus}\label{human-accuracy-across-participants-for-each-stimulus}

We will analyze accuracy for each stimulus by computing the proportion
of correct responses across all participants who viewed that stimulus.

\paragraph{Human accuracy across stimuli for each
participant}\label{human-accuracy-across-stimuli-for-each-participant}

We will analyze accuracy for each participant by computing the
proportion of correct responses across all stimuli.

\paragraph{Human-human consistency for each
stimulus}\label{human-human-consistency-for-each-stimulus}

We will estimate human-human consistency for each stimulus by computing
the proportion of responses that match the modal response for that
stimulus (whether that modal response is correct or incorrect).

\paragraph{Human-human consistency across stimuli (within
scenario)}\label{human-human-consistency-across-stimuli-within-scenario}

We will analyze human-human consistency by computing the mean
correlation between (binary) response vectors produced by each human
participant across all stimuli within each scenario.

\paragraph{Human accuracy as a function of stimulus
attributes}\label{human-accuracy-as-a-function-of-stimulus-attributes}

We will conduct exploratory analyses of human accuracy as a function of
various scenario-specific stimulus attributes that varied across trials.
We will examine those stimulus attributes that varied across stimuli
within each scenario and explore the relationship between each
individual attribute and human accuracy, as well as beetween linear
combinations of them and human accuracy.

\paragraph{Human accuracy by scenario}\label{human-accuracy-by-scenario}

We will fit human responses across all scenarios with a mixed-effects
logistic regression model, including scenario as a fixed effect and
participants and individual stimuli as random effects.

\paragraph{Other exploratory human behavioral
analyses}\label{other-exploratory-human-behavioral-analyses}

\begin{itemize}
\itemsep1pt\parskip0pt\parsep0pt
\item
  We will explore the relation of demographic variables on the
  performance of participants: how does age, gender, educational status
  and the the result of a one-trial spatial reasoning task relate to the
  overall accuracy of a subject?
\item
  We will additionally explore any potential left/right or yes/no
  response biases.
\end{itemize}

\paragraph{Human-model comparisons}\label{human-model-comparisons}

We will compare human and model behavior in two ways: \textbf{absolute
performance} and \textbf{response pattern.}

\subparagraph{\textbf{Absolute Performance}}\label{absolute-performance}

We will compare the accuracy of each model to the mean accuracy of
humans, for each scenario. To do this, we will first compute estimates
of mean human accuracy for each scenario and construct 95\% confidence
intervals for each of these estimates. These confidence intervals will
be constructed by bootstrapping: specifically, for an experiment with N
participants, we will resample N participants with replacement and
compute the proportion correct for that bootstrapped sample. We will
take repeat this resampling procedure 1000 times to generate a sampling
distribution for the mean proportion correct. The 2.5th and 97.5th
percentile will be extracted from this sampling distribution to provide
the lower and upper bounds of the 95\% confidence interval.

For each model, we will then compare their proportion correct (a point
estimate) to the human confidence interval.

\subparagraph{\textbf{Response Pattern}}\label{response-pattern}

We will compare the pattern of predictions generated by each model to
the pattern of predictions generated by humans.

We will do this by using two standard inter-rater reliability metrics:

\subparagraph{\textbf{Correlation between average-human and model
responses}}
For each stimulus, we will compute the proportion of ``hit'' responses
by humans. For each stimulus, we will extract the hit probability
generated by models. For each scenario (i.e., domain), we will compute
the root-mean-squared deviation between the human proportion-hit vector
and the model probability-hit vector. To estimate variability across
human samples, we will conduct bootstrap resampling (i.e., resampling
data from individual participants with replacement), where for each
bootstrap sample we will re-compute the correlation between the model
probability-hit vector and the (bootstrapped) human proportion-hit
vector.

\textbf{Cohen's kappa}

For each pair of human participants, we will compute Cohen's kappa
between their responses across the 150 stimuli, yielding a distribution
of pairwise human-human Cohen's kappa. The mutually exclusive categories
used in calculating Cohen's kappa is whether each of the 150 responses
was predicted to be positive or negative. For each model, we will
compute Cohen's kappa between its response vector and every human
participant, as well as every other model. A model's response pattern
will be considered more similar to humans' insofar as the mean
model-human Cohen's kappa (across humans) lies closer to the mean
human-human Cohen's kappa (for all pairs of humans).